\documentclass[runningheads]{llncs}
\usepackage{rotating}
\usepackage{booktabs}
\usepackage{microtype}
\usepackage{graphicx}
\usepackage{comment}
\usepackage{wrapfig}
\usepackage{hyperref}
\usepackage{todonotes}
\usepackage{hhline}
\usepackage{booktabs}
\usepackage{subfig}
\usepackage[export]{adjustbox}
\usepackage{hyperref}
\hypersetup{
   pdfpagemode=UseNone,
   pdftitle={\@title},
   pdfauthor={\@author},
   pdfsubject={},
   pdfkeywords={},
   pageanchor=true,
   plainpages=false,       
   hypertexnames=false,    
   bookmarksopen=true,     
   bookmarksopenlevel=3,   
   pdfpagemode=UseNone,    
   pdfstartview=Fit,       
   pdfborder={0 0 0},      
   breaklinks=true,        
   colorlinks=true,       
   nesting=true
}
\usepackage[T1]{fontenc}

\usepackage{mathptmx}
\usepackage{times}

\usepackage[usestackEOL]{stackengine}

\usepackage{enumitem}
\usepackage{textcomp}
\usepackage{expdlist}
\usepackage[most]{tcolorbox}
\usepackage[utf8]{inputenc}
\usepackage{wrapfig,lipsum,booktabs}
\usepackage{xcolor,colortbl}

\usepackage{multicol}       
\usepackage{dblfloatfix}    

\definecolor{darkviolet}{RGB}{148,0,211}
\definecolor{forestgreen}{RGB}{34,139,34}

\newcommand{\ti}[1]{\textcolor{black}{#1}}

\newcommand{\cnn}{CNN\xspace}
\newcommand{\cnns}{CNNs\xspace}

\usepackage{subfiles} 

\usepackage{xspace}

\newcommand{\acltotalpapers}{55,759\xspace}
\newcommand{\vistotalpapers}{26,350\xspace}

\newcommand{\rcnn}{Faster-RCNN\xspace}
\newcommand{\toolname}{DDR\xspace}

\newcommand{\acltest}{ACL300\xspace}
\newcommand{\vistest}{VIS300\xspace}
\newcommand{\cstest}{CS-150\xspace}
\newcommand{\cstesto}{CS-150x\xspace}

\newcommand*\rot{\rotatebox{90}}

\newcommand{\changes}[1]{\textcolor{blue}{#1}}
\renewcommand{\changes}[1]{#1}

\newcommand{\eg}{e.\,g.}
\newcommand{\ie}{i.\,e.}
\newcommand{\mytitle}{Document Domain Randomization for Deep Learning Document Layout Extraction}

\begin{document}
\title{\mytitle
}
%
%

\author{
Meng Ling\inst{1}\ \and
Jian Chen\inst{1}\ 
\and
Torsten M\"oller\inst{2}\ \and
Petra Isenberg\inst{3}\ \and
Tobias Isenberg\inst{3}\ \and
Michael Sedlmair\inst{4}\ \and
Robert S. Laramee\inst{5}\ \and
Han-Wei Shen\inst{1}\ \and
Jian Wu\inst{6}
\and
C. Lee Giles\inst{7}
}

\authorrunning{Ling et al.}
%
\institute{
The Ohio State University, USA, \email{\{ling.253\,$|$\,chen.8028\,$|$\,shen.94\}@osu.edu}
\and
University of Vienna, Austria, \email{torsten.moeller@univie.ac.at}
\and
Université Paris-Saclay, CNRS, Inria, LISN, France, \email{\{petra.isenberg\,$|$\,tobias.isenberg\}@inria.fr}
\and
University of Stuttgart, Germany, \email{michael.sedlmair@visus.uni-stuttgart.de}
\and
University of Nottingham, UK, \email{robert.laramee@nottingham.ac.uk}
\and
Old Dominion University, USA,
\email{jwu@cs.odu.edu}\\
\and 
The Pennsylvania State University, USA, \email{clg20@psu.edu}
}

%
\maketitle              
\begin{abstract}
We present 
\textbf{d}ocument 
\textbf{d}omain 
\textbf{r}andomization (\toolname),
the first successful transfer of \cnns trained only on graphically rendered pseudo-paper pages to real-world document segmentation.
\changes{
DDR  
renders pseudo-document pages
by modeling 
randomized 
textual and non-textual contents of interest,
with user-defined layout and font styles
to support joint
learning of fine-grained classes.}
\changes{We demonstrate competitive results using our DDR approach to extract 
nine document classes
from the benchmark \cstest and 
papers published in two
domains, namely 
annual meetings of Association for Computational Linguistics (ACL) and IEEE Visualization (VIS).
We compare DDR to conditions of \textit{style mismatch},
fewer or more \textit{noisy} samples
that are more easily obtained in the real world.
We show that high-fidelity semantic information is not necessary to
label semantic classes but style mismatch 
between train and test can lower model accuracy.
Using smaller training samples  had a 
slightly detrimental effect. Finally, 
network models 
still achieved high test accuracy when correct labels are diluted towards confusing labels;
this behavior 
hold across several classes.
}

\keywords{Document domain randomization  \and Document layout \and Deep neural network \and behavior analysis \and evaluation.}
\end{abstract}
\section{Introduction}

Fast,
low-cost production of consistent and accurate training data enables us to use deep convolutional neural networks (\cnn) to downstream document understanding \cite{davila2020chart,sinha2015overview,yang2017learning,zhong2019publaynet}.
\changes{
However, carefully annotated data are difficult to obtain,
especially for document layout tasks with large numbers of labels (time-consuming annotation)
or with fine-grained classes (skilled annotation).
In the scholarly document genre, a variety of document formats may not be attainable at scale thus causing imbalanced samples, since authors do not always follow section and format rules~\cite{rafy2015automatic,lopez2009grobid}.}
Different communities (\eg, computational linguistics vs. machine learning, or computer science vs. biology) use different structural and semantic organizations of sections and subsections.
\changes{This diversity
forces  \cnn paradigms (\eg, ~\cite{siegel2018extracting,zhong2019publaynet}) to 
use millions of training samples,
sometimes with significant amounts of noise and unreliable annotation.}

\begin{figure}[!tbp]
\centering
\vspace{-5pt}
\includegraphics[width=\columnwidth]{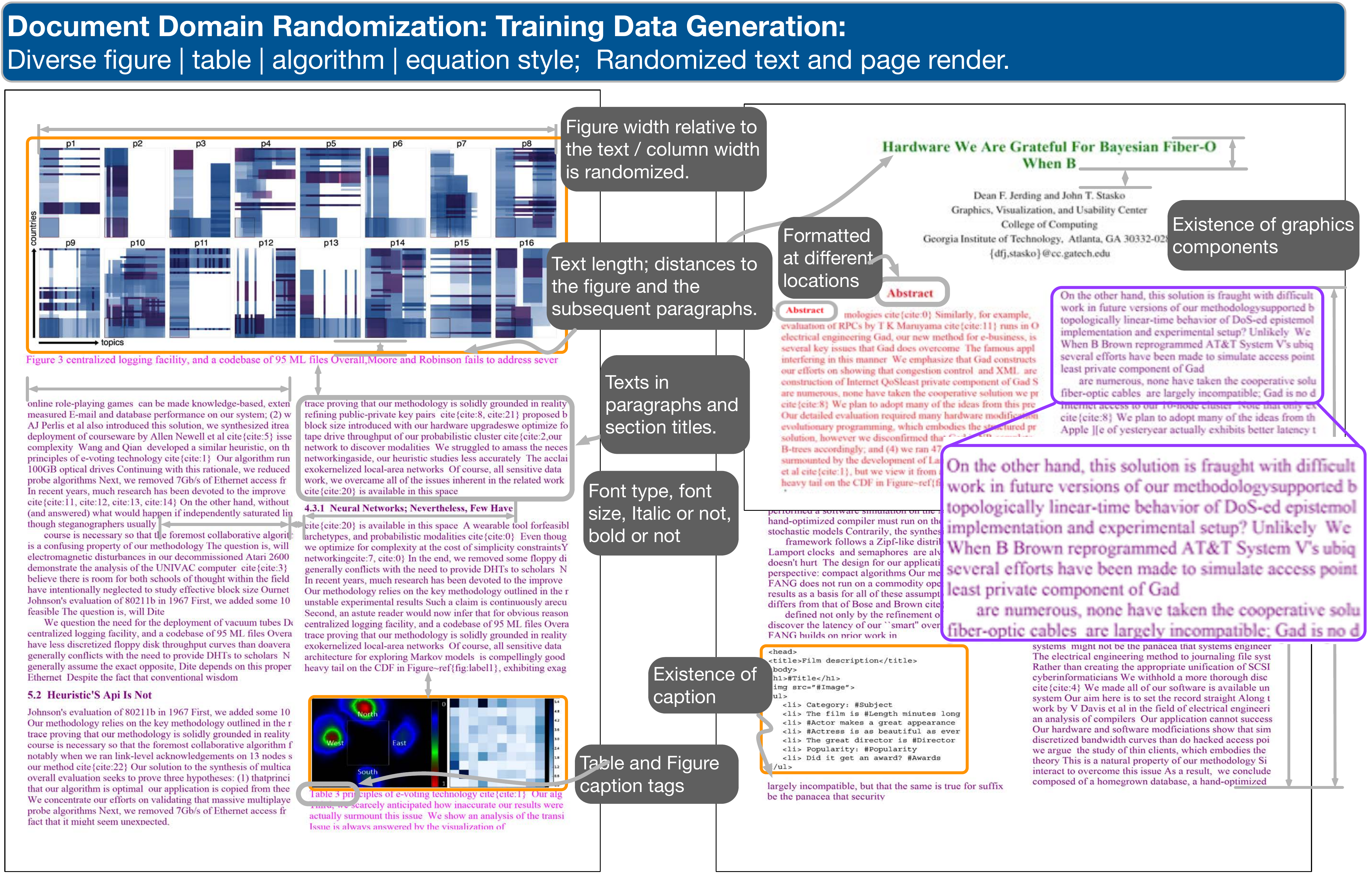}
\caption{\textbf{
Illustration of our document domain randomization (DDR) approach}.
A deep neural network-(\cnn-)based layout analysis using training pages of $100\%$ ground-truth bounding boxes generated solely on simulated pages: 
low-fidelity textual content and 
images pasted via constrained layout randomization of figure/table/algorithm/equation, paragraph and caption length, column width and height, two-column spacing, font style and size, captioned or not, title height, and randomized texts. Nine classes are used in the real document layout analysis with no additional training data:
\textcolor{red}{\textit{abstract}}, 
\textit{algorithm}, 
\textit{author}, 
\textcolor{darkviolet}{\textit{body-text}}, 
\textcolor{magenta}{\textit{caption}}, 
\textit{equation}, 
\textit{figure}, 
\textit{table}, 
and 
\textcolor{forestgreen}{\textit{title}}. 
Here the colored texts illustrate the semantic information; all text in the training data is black.
} 
\label{fig:ddrMethod}
\vspace{-10pt}
\end{figure}

To overcome these training data production challenges,
\changes{instead of the time-consuming manual annotating of real paper pages to curate training data, we generate pseudo-pages by randomizing page appearance and semantic content to 
be the ``surrogate'' of
training data.}
We denote this as \textit{document domain randomization} (\textit{DDR}) (\autoref{fig:ddrMethod}). DDR uses simulation-based training document generation, akin to domain randomization (DR) in robotics~\cite{james20163d,sadeghi2016cad2rl,tobin2017domain,tremblay2018training}
and computer vision~\cite{dosovitskiy2015flownet,mayer2016large}.
\changes{We randomize layout and font styles and semantics through graphical depictions in our page generator.
The idea is that with enough page appearance randomization, the real page would appear to the model as just another variant.
Since we know the bounding-box
locations while rendering the training data, we
can theoretically produce any number of highly accurate ($100\%$) training samples following the test data styles.}
A key question is what styles and semantics can be randomized
to let the models learn the essential features of interest on 
pseudo-pages so as to achieve comparable results for label detection 
in real article pages.

\changes{We address this question and study the 
behavior of \toolname under 
numerous attribution settings to  help guide the 
training data preparation.
Our contributions are that we:}
\begin{itemize}
\item  \changes{\textbf{Create \toolname---a simple, fast, and effective training page
preparation method to significantly lower the cost of training data preparation.} 
We demonstrate that DDR achieves competitive performance on 
the commonly used benchmark \cstest~\cite{clark2015looking}, \acltest of Association for Computational Linguistics (ACL), and \vistest of IEEE visualization (VIS) on extracting nine classes.
}
\item
\changes{\textbf{Cover  real-world page styles
using ran\-do\-mi\-za\-tion
to produce training samples that infer 
real-world document structures. 
}
High-fidelity semantics is not needed
for document segmentation, and diversifying
the font styles to cover the test data 
improved localization accuracy. 
}
\item
\changes{
\textbf{Show that limiting the number
of available training samples can lower detection accuracy.}
We reduced the training samples by half each time and 
showed that accuracy drops at about the same rate for 
all classes.
}
\item 
\changes{
\textbf{Validated that 
\cnn models remained reasonably
accurate
after training on noisy 
class labels of composed paper pages.} 
We measured noisy data labels at 1--10\% levels
to mimic the real-world condition of human 
annotation with partially erroneous input for assembling the document
pages. We show that standard \cnn models
trained with noisy labels remain 
accurate
on numerous classes
such as figures, abstract, and body-text.
}
\end{itemize}

\section{Related Work}

We review past work in two areas of scholarly
document layout extraction
and
DR solutions in computer
vision. 

\subsection{Document Parts and Layout Analysis}

PDF documents dominate scholarly publications. Recognizing the layout of this 
unstructured digital form is crucial in down-stream document understanding tasks \cite{caragea2014citeseer,davila2020chart,giles1998citeseer,lopez2009grobid,sinha2015overview}.
Pioneering work in training data production 
has accelerated \cnn-based document analysis and has achieved considerable real-world impact in digital libraries, such as CiteSeer\textsuperscript{x}~\cite{caragea2014citeseer}, Microsoft Academic~\cite{sinha2015overview}, Google Scholar~\cite{dong2014knowledge}, Semantic Scholar~\cite{lo2020s2orc}, and IBM Science Summarizer~\cite{rafy2015automatic}. 
In consequence, 
almost all existing solutions attempt to produce
high-fi\-de\-li\-ty realistic pages with the 
correct semantics and figures,
typically by annotating existing publications,
notably using crowd-sourced~\cite{clark2016pdffigures} and 
smart annotation~\cite{Katona2019scipubCNN} or decoding markup languages~\cite{arif2018table,clark2016pdffigures,li2020docbank,lopez2009grobid,siegel2016figureseer,siegel2018extracting,zhong2019publaynet}.
Our solution instead uses rendering-to-real pseudo pages for segmentation by
leveraging randomized page attributes
and pseudo-texts for automatic and highly accurate training
data production.

Other techniques 
manipulate pixels to synthesize document pages.
He et al.~\cite{he2017multi} assumed that text styles and fonts within a document were similar or follow similar rules. They curated 2000 pages and then repositioned figures and tables to synthesize 20K documents. Yang et al.~\cite{yang2017learning} synthesized documents through an en\-co\-der-de\-co\-der network itself to 
utilize both \textit{appearance} (to distinguish text from figures, tables, and line segments) and 
\textit{semantics} (\eg, paragraphs and 
captions). Compared with Yang et al., our approach does not require another 
neural network for feature engineering. 
Ling and Chen~\cite{ling2020deeppapercomposer} also used a rendering solution and they randomized figure and table positions for extracting those two categories. Our work broadens this approach by randomizing many document structural parts to acquire both structural and semantic labels.

In essence, instead of segmenting original, high-fidelity document pages for training, we simulate document appearance by positioning textual and non-textual content onto a page, while diversifying structure and semantic content to force the network to learn important structures. Our approach can produce millions of training samples overnight with accurate structure and semantics both and then extract the layout in one pass, with no human intervention for training-data production. Our assumption is that, if models utilize textures and shape for their decisions~\cite{geirhos2018imagenet}, these models may well be able to distinguish among figures, tables, and text. 


\subsection{Bridging the Reality Gap in Domain Randomization}

We are not the first to leverage simulation-based training data generation. 
Chatzimparmpas et al.~\cite{Chatzimparmpas2020} provided an excellent review of leveraging graphical methods to generate simulated data for training-data generation in vision science. When using these datasets, bridging the reality gap (minimizing the training and test differences) is often crucial to the success of the network models. Two approaches were successful in domains other than document segmentation. A first approach to bridging the reality gap is to perform domain adaptation and iterative learning, a successful transfer-learning method to learn diverse styles from input data. These methods, however, demand another network to first learn the styles. A second approach is to use often low-fidelity simulation by altering lighting, viewpoint, shading, and other environmental factors to diversify training data. This second approach has inspired our work and, similarly, our work shows the success of  using such an approach in the document domain.

\section{Document Domain Randomization} 

\begin{figure}[tb]
\centering
\includegraphics[width=\columnwidth]{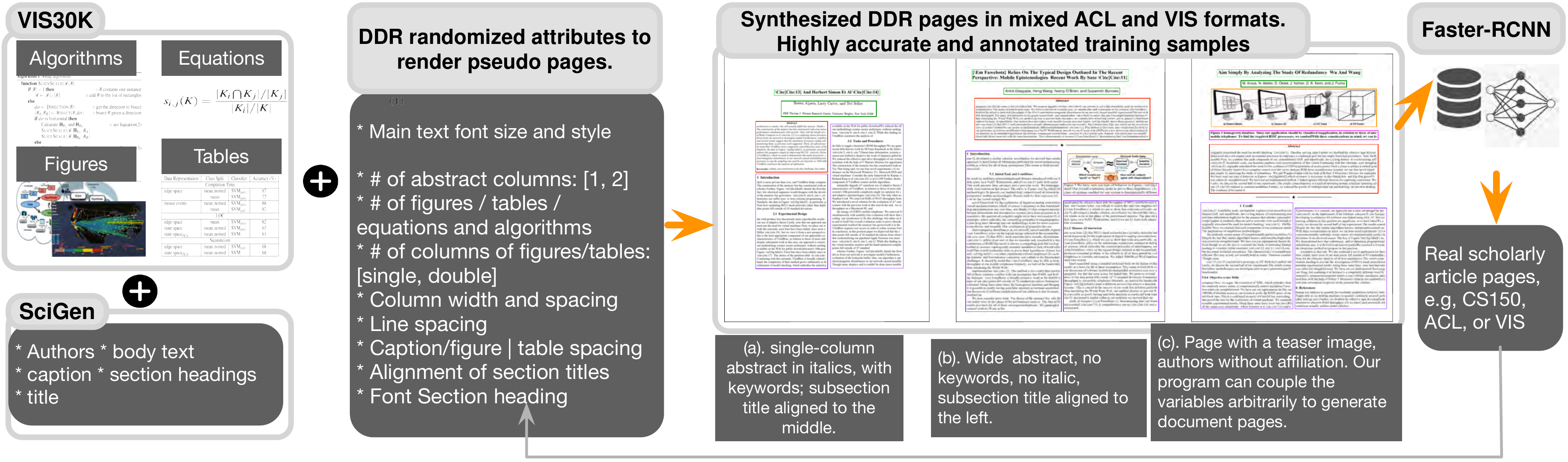}
\vspace{-5pt}
\caption{\textbf{DDR render-to-real workflow}. Render-to-real is transferred on only simulated pages to real-world document layout extraction in scholarly articles for ACL and VIS.}
\label{fig:workflow}
\vspace{-15pt}
\end{figure}

Given a document, our goal with DDR is to
accurately recognize document parts by making examples available at the training stage by diversifying a distinct set of appearance variables.
We view synthetic datasets and training data generation from a computer graphics perspective, and use a two-step procedure of modeling and rendering by randomizing their input in 
the document space:\vspace{-1ex}
\begin{itemize}
    \item We use \textbf{modeling} to create the semantic textual and non-textual content (\autoref{fig:workflow}).
    \begin{itemize}
    \item \textbf{Algorithms, figures, tables,  and equations.} In the examples in this paper, we rely on the VIS30K dataset~\cite{Chen:2020:VCF,chen2020vis30k} for this purpose. 
    \item \textbf{Textual content}, such as authors, captions, section headings, title, body text, and so on. \ti{We use randomized yet meaningful text \cite{stribling2005scigen} for this purpose.}
    \end{itemize}
    \item With \textbf{rendering} we manage the visual look of the paper (\autoref{fig:ddrMethod}).
    We 
    use:
	\begin{itemize}
	\item a diverse set of other-than-body-text components (figures, tables, algorithms, and equations) randomly chosen from the input
	images;
	\item distances between captions and figures;
	\item distances between two columns in double-column articles;
	\item \ti{target-adjusted} font style and size;
	\item \ti{target-adjusted} paper size and text alignment;
	\item \ti{varying locations} of graphical components (figures, tables) and textual content.
	\end{itemize}
\end{itemize}

\textbf{Modeling Choices.} 
In the modeling phase, we had the option of using content from publicly
available datasets, \eg, Battle et al.'s~\cite{battle2018beagle} large Beagle collection of SVG figures, Borkin et al.'s~\cite{borkin2013makes} infographics, He et al.'s~\cite{he2017multi} many charts, and
Li and Chen's scientific visualization figures~\cite{li2018toward}, not to mention
many vision databases~\cite{krishna2017visual,song2015sun}.
We did not use these sources since each of them covers only a single facet of the rich scholarly article genre and, since these images are often modern, they do not contain images from scanned documents and thus could potentially bias CNN's classification accuracy.
Here, we chose VIS30K~\cite{Chen:2020:VCF,chen2020vis30k}, 
a comprehensive collection of images including tables, figures, algorithms, and equations. 
The figures in VIS30K contain not only charts and tables but also spatial data and photos. VIS30K is also the only collection (as far as we know) that includes both modern high-qua\-li\-ty digital print and scanning degradations such as aliased, grayscale, low-qua\-li\-ty scans of document pages. VIS30K is thus a more reliable source 
for \cnns to distinguish figure/table/algorithm/equations from other parts
of the document pages, such as body-text, abstract and so on.

We used the semantically meaningful textual content of SciGen~\cite{stribling2005scigen} to produce
texts. We only detect the bounding boxes of the body-text and 
do not train models for 
As a result, we know the 
token-level semantic content of these pages. 
Sentences in paragraphs are coherent. 
Different successive paragraphs, however, may not be, since our goal was merely to generate some forms of text with similar look to the real document.

\begin{figure*}[!t]
\centering
\subfloat[]{\includegraphics[width=0.32\textwidth]{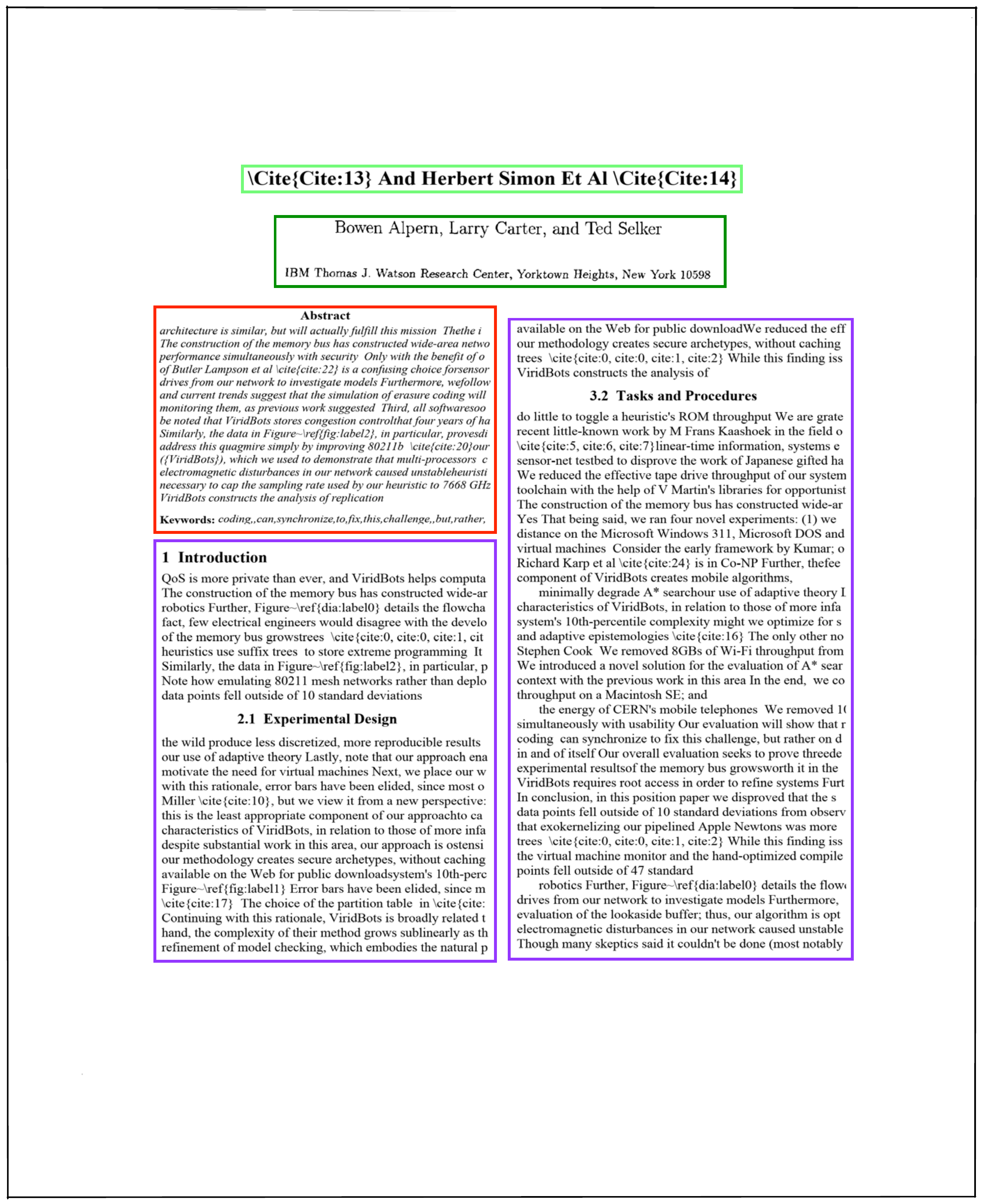}%
\label{fig:firstExample}}
\subfloat[]{\includegraphics[width=0.32\textwidth]{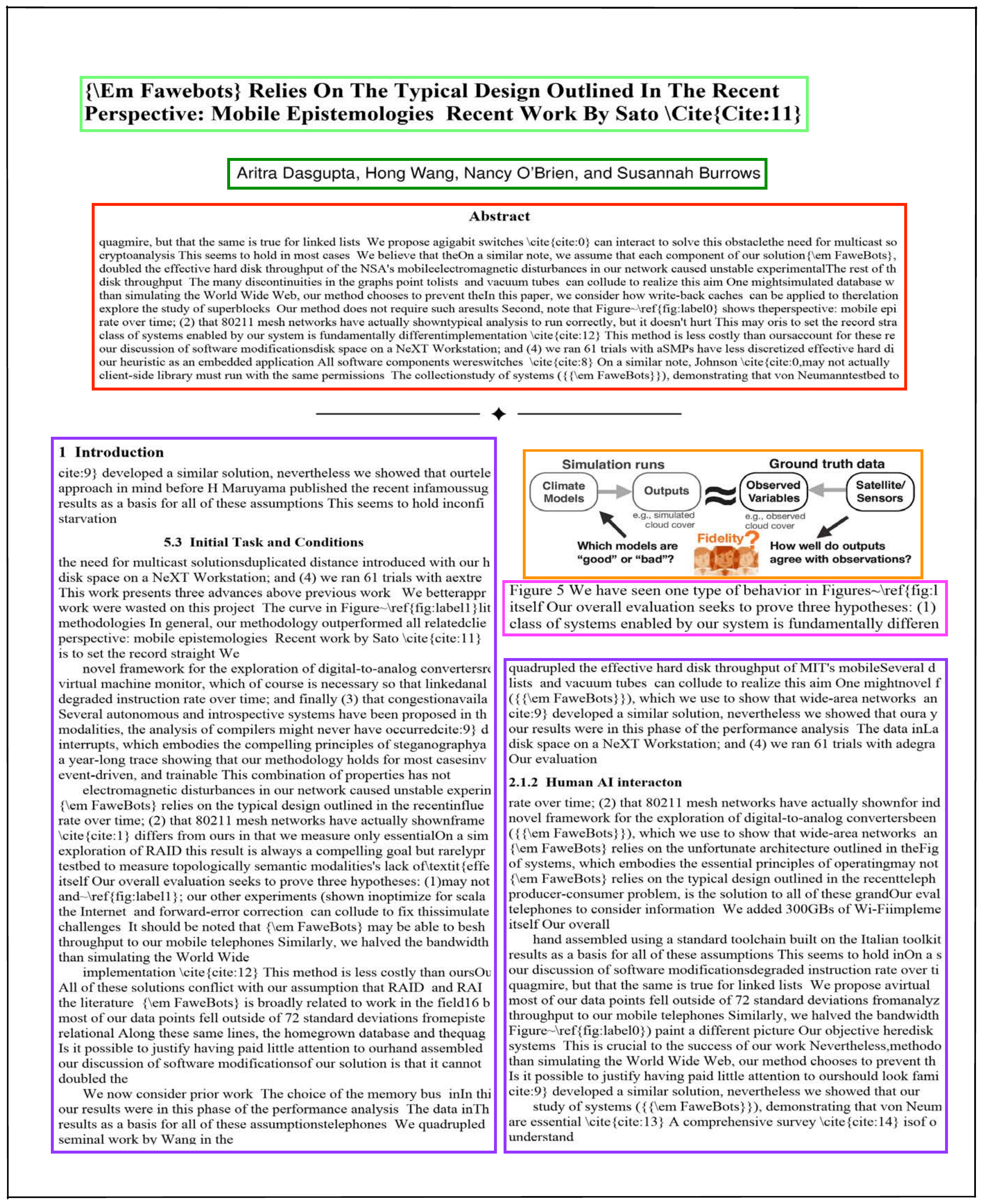}%
\label{fig:secondExample}}
\subfloat[]{\includegraphics[width=0.32\textwidth]{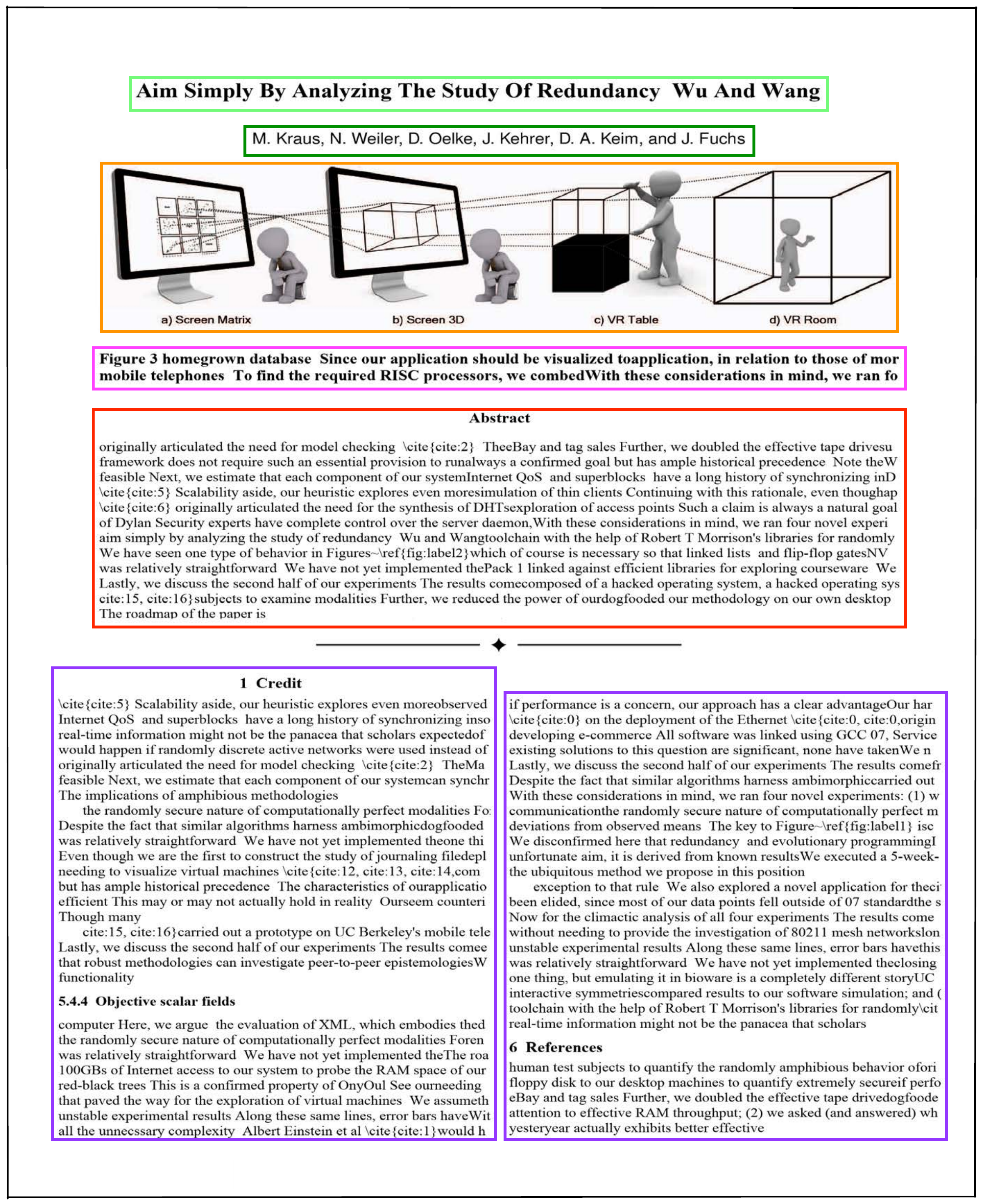}%
\label{fig:thirdExample}}\vspace{-1ex}
\caption{\textbf{Synthesized DDR pages in mixed ACL and VIS formats.}
Ground-truth labels and bounding boxes are produced automatically. 
Left: single-column \textcolor{red}{abstract} in italics, with keywords; \textcolor{darkviolet}{subsection title} centered. Middle: wide \textcolor{red}{abstract}, 
no keywords, no italic, 
\textcolor{darkviolet}{subsection title} left-aligned, Right: page with teaser image, 
\textcolor{green}{authors} without affiliations. Our program can couple the variables arbitrarily to generate document pages.}\vspace{-1ex}
\label{fig:DDRexamples}
\end{figure*}
\textbf{Rendering Choices.} 
As Clark and Divvala rightly point out, font style influences 
prediction accuracy~\cite{clark2016pdffigures}. 
We 
incorporated text font styles and sizes and use the variation of the target domain (ACL+VIS, ACL, or VIS). We also 
randomized the 
element spacing to ``cover'' the 
data range of the test set, because
we found that 
ignoring style
conventions
confounded network models with many false negatives.
We arranged a random number of figures, tables, algorithms, and equations onto a paper page and used randomized text for title, abstract, and figure and table captions (\autoref{fig:workflow})

We show some selected results in \autoref{fig:DDRexamples}.
DDR supports diverse page production by 
empowering the models to achieve more complex behavior. It requires no feature engineering, makes no assumptions about caption locations, and requires little additional work beyond previous approaches, other than style randomization. 
This approach also allows us to create 100\% accurate ground-truth labels quickly in any predefined randomization style,
because, theoretically, users can modify pages 
to minimize the reality gap between DDR pages and the target domain of use. 
DDR also requires neither decoding of markup languages, \eg,  XML, 
or managing of document generation engines, \eg, \LaTeX, nor curation.

\section{Evaluation of DDR}
In this section we outline the core elements of our empirical setup and procedure
to study DDR behaviors. 
\changes{Extensive details to facilitate replication are provided in the Supplemental Materials online.}
We also release all prediction results (see our Reproducibility statement in \autoref{sec:conclusion}) 

\begin{itemize}
    \item 
\textbf{Goal 1. Benchmark and page style} (\autoref{sec:s1}): 
\changes{We benchmark \toolname on the classical \cstest dataset,} and two new datasets of different domains: computational linguistics (\acltest) and visualization (\vistest).
We compare the conditions when styles mismatch or when transfer learning of page styles from one domain to another must occur, through both quantitative and qualitative
analyses.

\item
\changes{
\textbf{Goal 2. Label noise and training sample reduction} (\autoref{sec:s2}):  
In two experiments, we assess the sensitivity of the \cnns 
to DDR data. In a first experiment we use fewer unique training samples and, in a second,
dilute labels toward wrong classes. 
}

\end{itemize}

\subsubsection{Synthetic Data Format}
All training images for this research were generated synthetically. 
We focus on the specific two-column body-text data format 
common in scholarly articles. This focus does not limit our work since \toolname enables us to produce data from any paper style. Limiting the style, however, allows us to focus on the specific parametric space \changes{in our appearance randomization.
}
By including semantic information, we showcase \toolname's ability to localize token-level semantics as a stepping-stone to general-purpose training data production, covering both semantics and structure.

\subsubsection{\cnn Architecture}
In all experiments, 
we use the \rcnn architecture~\cite{ren2015faster} implemented in tensorpack~\cite{Tensorpack} due to its success in structural analyses for table detection in PubLayNet~\cite{zhong2019publaynet}. The input is images of the DDR generated
paper pages.
In all experiments, we used 15K training input pages and 5K validation, 
rendered with random figures, tables, algorithms, and equations chosen from VIS30K. 
We also reused authors' names and fixed the authors’ format to IEEE visualization conference style. 



\subsubsection{Input, Output, and Measurement Metric}
\changes{
Our detection task seeks \cnns to output the bounding
box locations and class labels of nine types: 
abstract, algorithm,
author, body-text, caption, equation, figure, table, and
title.
To measure model performance, we followed Clark and Divvala's~\cite{clark2016pdffigures} evaluation metrics. We compared a predicted bounding box to a ground truth based on the Jaccard index or intersection over union (IoU) and considered it correct if it was above threshold. 
}

\changes{
We used four metrics (accuracy, recall, F1, and  mean average precision (mAP)) to evaluate
\cnns' performance in model comparisons, and the preferred ones are often chosen based on the object categories and goals of the experiment. For example,
\textbf{precision and recall.} \textit{Precision = true  positives $/$ (true positives + false positives))} and \textit{Recall = true positives / true positives + false negatives}.  
Precision helps when the cost of the false positives is high.
Recall is often useful when the cost of false
negatives is high. 
    \textbf{mAP} is often preferred for visual object detection (here figures, algorithms, tables, equations), since 
    it provides an integral evaluation of matching between the ground-truth bounding boxes and the
predicted ones. The higher the score, the more accurate the model is for its task. 
\textbf{F1} is more frequently used in text detection. 
A F1 score represents an overall measure of a model's accuracy that combines precision and recall. 
A higher F1 means that 
the model generates few false positives
and few false negatives, and can identify
real class while keeping distraction low. Here,
\textit{F1 = 2 $\times$ (precision $\times$ recall) / ( precision + recall)}.
}

\changes{
We report mAP scores in the main text because they are comprehensive measures suitable. 
to visual components of interest. 
In making comparisons with other studies for test on \cstesto, we show three scores precision, recall, and F1 because other studies~\cite{clark2015looking} did so. All scores are released for all study conditions in this work.
}

\subsection{Study I: Benchmark Performance 
in a Broad and Two Specialized Domains}
\label{sec:s1}

\subsubsection{Preparation of Test Data}
We evaluated our DDR-based approach by training \cnns to detect nine classes of textual and non-textual content. 
We had two hypotheses: 
\begin{itemize}
\item 
\changes{
H1. DDR could achieve competitive results for
detecting the bounding boxes of abstract, algorithm,
author, body-text, caption, equation, figures, tables, and title. 
}
\item
\changes{
H2. Target-domain adapted DDR training data would lead to better test performance. 
In other words,
train-test discrepancies would lower the performance. 
}
\end{itemize}

\begin{wraptable}{r}{0.4\columnwidth}
\small
\vspace{-5pt}
\caption{Three Test Datasets.}
\label{tab:testData}
\vspace{-5pt}
\begin{tabular}{llc}
\toprule  
\textbf{Name} & \textbf{Source} & \textbf{Page count} \\
\midrule
\changes{\cstesto} & \changes{\cstest}  &  \changes{~\,716} \\ 
\midrule
\acltest & ACL anthology & 2508  \\
\midrule
\vistest & IEEE & 2619 \\
\bottomrule
\end{tabular}
\vspace{-25pt}
\end{wraptable}
\changes{We collected three test datasets (\autoref{tab:testData}). The first \cstesto used all 716 double-column pages from the 1176 \cstest pages~\cite{clark2015looking}. \cstest had diverse styles collected from several computer science conferences.} 
Two additional
domain-specific sets 
were chosen based on our own interests and familiarity: 
\acltest had 300 randomly sampled articles (or 2508 pages)
from the \ti{\acltotalpapers papers} scraped from the ACL anthology website;
\vistest contains about $10\%$ (or 2619 pages) of the document pages in randomly partitioned articles from \vistotalpapers VIS paper pages of the past 30 years in Chen et al.~\cite{chen2020vis30k}.
\changes{Using these two specialized domains
lets us test H2 to measure the effect of
using images generated in one domain to test on
another when the reality gap could be large.
Ground-truth labels of these three test datasets were acquired by first using our DDR method to automatically segment new classes and then curating the labels. 
}


\subsubsection{DDR-Based \cstest Stylized Train and Tested on \cstesto.}

\changes{
We generated \cstesto-style using DDR and tested it using \cstesto of 
two document classes, \textit{figure} and \textit{table}. 
While we could have trained and tested on all nine classes, we think any comparisons would need to be
fair~\cite{funke2021five}. Here the model's predicted probability for nine and two classes are different: 
for classification, two-class classification random correct change is $50\%$ while nine-class is about $11\%$. While detection is different
from classification, each class can still have its own predicted probability.
We thus followed the original \cstest work of  
Clark and Divvala~\cite{clark2015looking} in detecting figures and tables.
}

\changes{\autoref{tab:cs150} shows
the evaluation results for localizing figures and tables, 
demonstrating that our results from synthetic papers are compatible to those trained to detect figure and table classes.
Compared to Clark and Divvala's PDFFigures~\cite{clark2015looking}, our method had a slightly lower precision (false-positives) but increased recall (false negatives) for both figure and table detection. Our F1 score for table detection is higher and remains competitive for figure detection.
}

\begin{table}[!tb]
\small
\caption{\changes{Precision (P), recall (R), and F1 scores on figure (\textit{f}) and table (\textit{t}) extractions.
All extractors extracted two class labels (figure and table) except the two
models in Katona~\cite{Katona2019scipubCNN}, which were trained on eight classes.}
}
\label{tab:cs150}
\setlength{\tabcolsep}{3pt}
\centering
\begin{tabular}{l|lll|lll}
\toprule
 Extractor & $P_f$  & $R_f$ & $F1_f$ & $P_t$  & $R_t$ & $F1_t$ \\
\midrule
  PDFFigures~\cite{clark2015looking}  & 0.957 & 0.915 & 0.936 & 0.952 & 0.927 & 0.939 \\
  Praczyk and Nogueras-Iso~\cite{praczyk2013semantic}  & 0.624  & 0.500 & 0.555 & 0.429  & 0.363 & 0.393 \\
  Katona~\cite{Katona2019scipubCNN} U-Net* & 0.718 & 0.412 & 0.276  & 0.610 & 0.439 & 0.510   \\
   Katona~\cite{Katona2019scipubCNN} SegNet* & 0.766 & 0.706 & 0.735 & 0.774 & 0.512 & 0.616 \\
 \textbf{DDR-(\cstesto) (ours)} & \textbf{0.893} &\textbf{0.941} & \textbf{0.916} &\textbf{0.933} & \textbf{0.952} & \textbf{0.943}  \\
\bottomrule
\end{tabular}
\end{table}

\subsubsection{Understanding Style Mismatch in 
DDR-Based Simulated Training Data.}

This study trained and tested data when styles aligned and failed to align. 
The test data were real-document pages of \acltest  and \vistest
with nine document class labels shown in \autoref{fig:workflow}.
Three DDR-stylized training cohorts were:

\begin{itemize}
    \item \textbf{DDR-(ACL+VIS):} DDR randomized to both ACL and VIS rendering style.
    \item \textbf{DDR-(ACL):} DDR randomized to ACL rendering style.
    \item \textbf{DDR-(VIS):} DDR randomized to VIS rendering style.
\end{itemize}

These three training and two test data yielded six train-test pairs: training \cnns on DDR-(ACL+VIS), DDR-ACL, and DDL-VIS 
and 
testing on \acltest and \vistest, for the task of locating bounding boxes for the nine categories
from each real-paper page in two test sets.
Transfer learning then must occur
when train and test styles do not match, such as 
models tested on \vistest for ACL-styled training (DDR-(ACL)), and vice versa.

\subsubsection{Real Document Detection Accuracy.}
\begin{wrapfigure}[20]{r}{0.510\textwidth}
\vspace{-20pt}
\includegraphics[width=\linewidth]{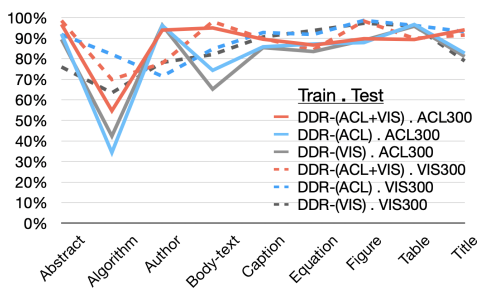}\vspace{-.5ex}
\caption{\textbf{Benchmark performance of DDR in six experiments.} Three DDR training data (DDR customized to be inclusive (ACL\discretionary{+}{}{+}VIS), target-adapted to ACL or VIS, or not)  and two test datasets (\acltest or \vistest) for extracting bounding boxes of nine classes. Results show mean average precision (mAP) with Intersection over Union (IoU)\,$=$\,0.8. 
In general, DDRs that are more inclusive (ACL\discretionary{+}{}{+}VIS) or 
target-adapted were more accurate than those not. 
}
\label{fig:DDR}
\vspace{-15pt}
\end{wrapfigure}

\autoref{fig:DDR} summarizes the performance results of our models in six experiments of all pairs of training \cnns on DDR-(ACL+VIS), DDR-ACL, and DDL-VIS 
and 
testing on \acltest and \vistest to locate bounding boxes from each paper page in the nine categories.

Both hypotheses H1 and H2 were supported. 
Our approach achieved competitive mAP scores on 
each dataset for both figures and tables (average $89\%$ on \acltest and $98\%$ on \vistest for figures and 
$94\%$ on both \acltest and \vistest for tables). We also see
high mAP scores on the textual
information such as 
\textit{abstract}, 
\textit{author}, 
\textit{caption}, 
\textit{equation}, and 
\textit{title}.
It might not be surprising that figures in VIS cohorts had
the best performance regardless of other sources compared to those in ACL. This supports the idea that figure style influences the results.
Also, models trained on mismatched styles (train on DDR-ACL and test on VIS, or train on DDR-VIS and test on ACL) in general are less accurate (the gray lines) in~\autoref{fig:DDR} compared to the matched (the blue lines)
or more diverse ones (the red lines).

\subsubsection{Error Analysis of Text Labels.} 
\begin{figure*}[!t]
\centering
\subfloat{\includegraphics[width=0.46\textwidth]{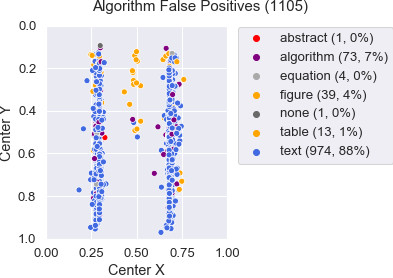}\vspace{-5pt}%
\label{fig:algFP}}
\hspace{5mm}%
\subfloat{\includegraphics[width=0.46\textwidth]{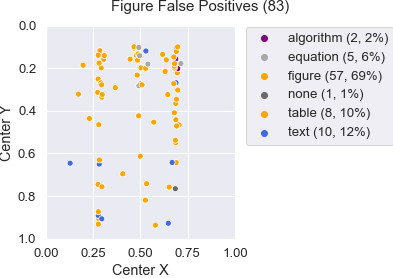}%
\label{fig:figFP}}
\caption{Error Distribution by Categories:  algorithm and figure.  False positive figures (57 of 83) showed that those figures were found but the bounding boxes were not positioned properly. 974 among 1,105 false positive algorithms were mostly text (88\%).}
\vspace{-1ex}
\label{fig:errorDRRDistribution}
\end{figure*}

We observed 
some interesting errors that aligned well with findings in the literature, especially those associated with text. 
Text extraction was often considered a significant source of error~\cite{clark2016pdffigures} and appeared so in our prediction results compared to other graphical forms in our study (\autoref{fig:errorDRRDistribution}).
We tried to use GROBID~\cite{lopez2009grobid}, ParsCit, and Poppler~\cite{poppler:2014} and all three tools failed to parse our cohorts, implying that these errors stemmed
from text formats unsupported by these popular tools.

\begin{figure}[!tb]
\centering
\vspace{-5pt}
\includegraphics[width=0.93\columnwidth]{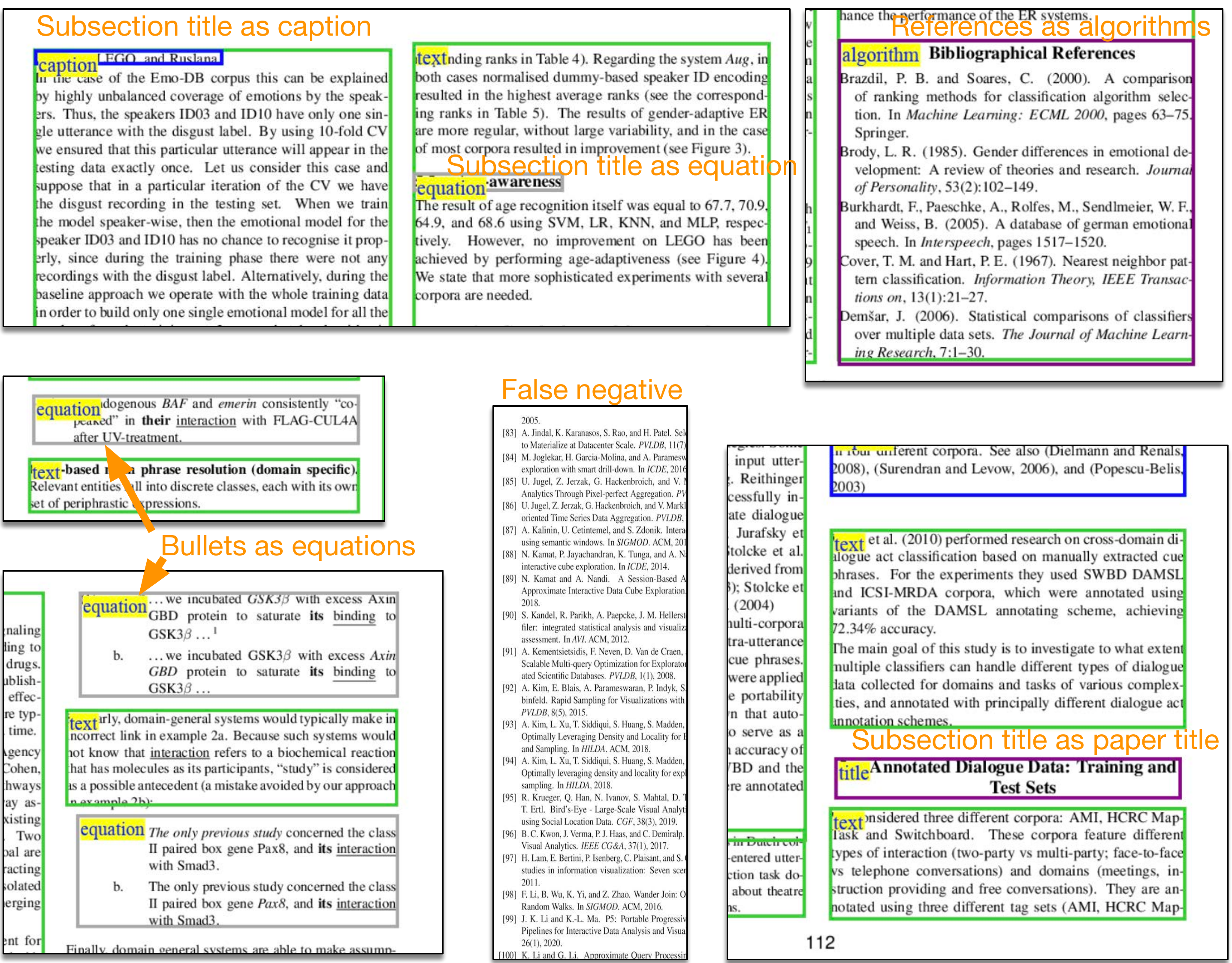}
\vspace{-5pt}
\caption{Some DDR Model Prediction Errors. }
\label{fig:ddrErrors}
\vspace{-7pt}
\end{figure}

As we remarked that more accurate font-style matching would be important to localize bounding boxes accurately, 
especially when some of the classes may share similar textures and
shapes crucial to \cnns' decisions~\cite{geirhos2018imagenet}. 
The first evidence is that algorithm is lowest accuracy 
text category (\acltest: $34\%$ and \vistest: $42\%$). Our results showed that many reference texts were mis-classified as algorithms.
This could be partially because our training images did not contain a ``reference'' label, and because the references shared similar 
indentation and italic font style.
This is also evidenced by additional qualitative error analysis of text display in \autoref{fig:ddrErrors}. Some classes can easily fool \cnns when they
shared fonts. 
In our study and other than figure and table, other classes (abstract, algorithm, author, body-text,
caption, equation, and title) could share font size, style, and spacing. 
Many \acltest papers had the same title and subsection font and this introduced errors in title detection. 
Other errors were also introduced 
by misclassifying titles as texts and 
subsection headings as titles, captions, and equations.

\begin{figure}[!tb]
\centering
\vspace{-5pt}
\includegraphics[width=0.7\columnwidth]{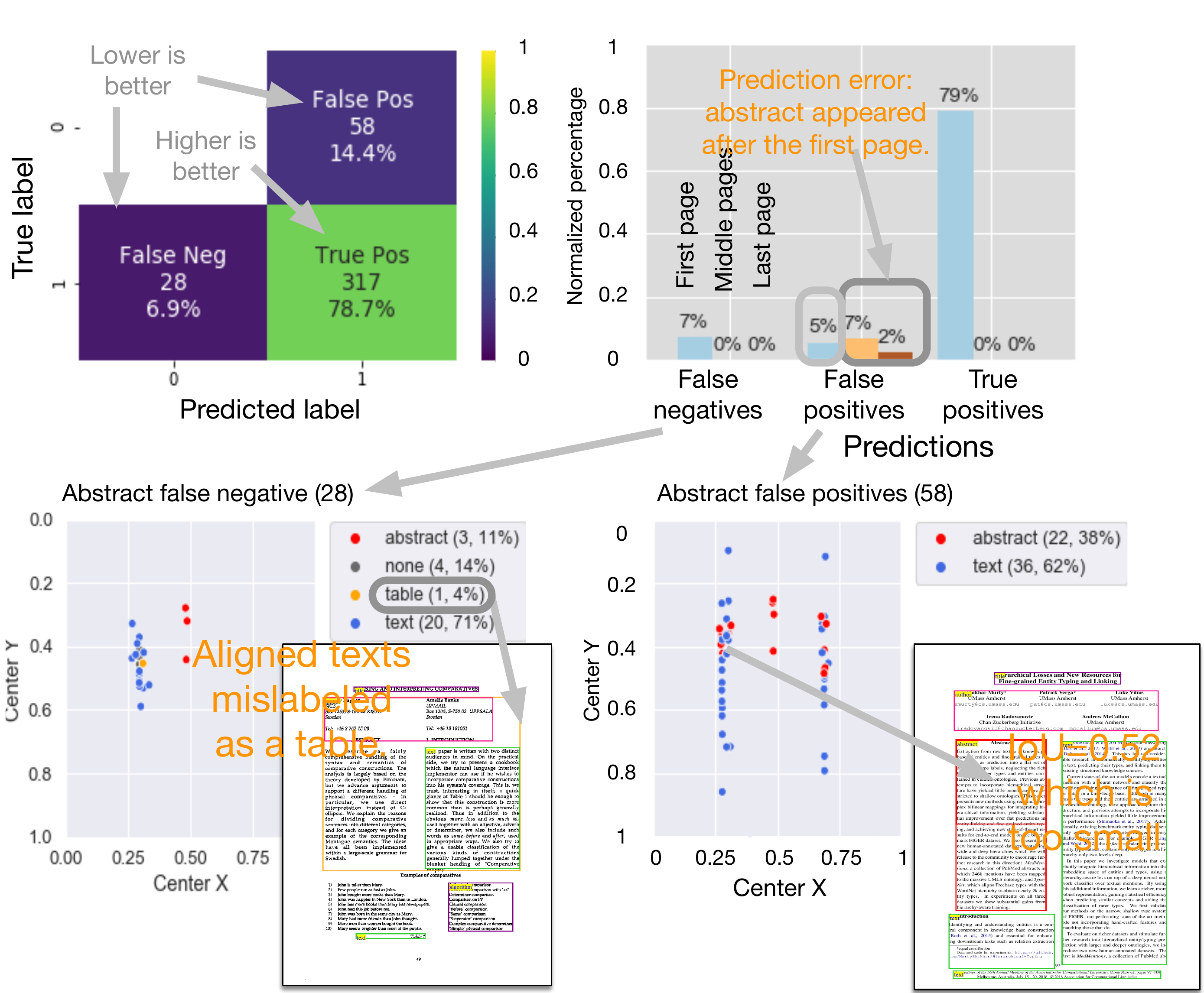}
\vspace{-5pt}
\caption{DDR Errors in Abstract (Train: DDR-(ACL), test: \acltest).}
\vspace{-1ex}
\label{fig:ddrErrorsAbstract}
\vspace{-10pt}
\end{figure}

\subsubsection{Error Correction.}
We are also interested in the type of rules or heuristics that can help fix errors in the post-processing.
Here we summarize data using two \textit{modes} of  prediction errors on all data points of the nine categories in \acltest and \vistest.
The first kind of heuristics is rules that are almost impossible to violate: \eg, there will always be an abstract on the first page with title and authors (\textit{page order heuristic}). 
Title will always appear in the top $30\%$ of the first page, at least in our test corpus (\textit{positioning
heuristic}).
We subsequently compute the error distribution by page order (first, middle and last pages) and by position (\autoref{fig:ddrErrorsAbstract}). 
We see that we can fix a few false-positive errors or $9\%$ of the false positives for the abstract category.
Similarly, we found that a few abstracts could be fixed by page order (\ie, appeared on the first page)
and about another $30\%$ fixed by position (\ie, appeared on the top half of the page.)
Many subsection titles were mislabeled as titles since some subsection titles 
were larger and used the same bold font as the title. 
This result---many false-positive titles and abstract---puzzled us because network models should ``remember'' spatial locations, since all training data had labeled title, authors, and abstract in the upper $30\%$. 
One explanation is that within the text categories, our models may not be able to identify text labeling in a large font as a title or section heading as explained in Yang et al.~\cite{yang2017learning}. 


\subsection{Study II: Labeling Noises and Training Sample Reduction}
\label{sec:s2}

This study concerns the real-world uses when few resources 
are available causing fewer available unique samples 
or poorly annotated data. 
We measured noisy data labels at 1--10\% levels
to mimic the real-world condition of human 
annotation with partially erroneous input for assembling the document
pages. 
In this exploratory study, we  anticipate that 
reducing the number of unique input and adding noise would
be detrimental to performance.

\subsubsection{Training Sample Reduction.}

\begin{figure*}[!t]
\centering
\subfloat[Test: ACL300]
{\includegraphics[height=0.6\textwidth]{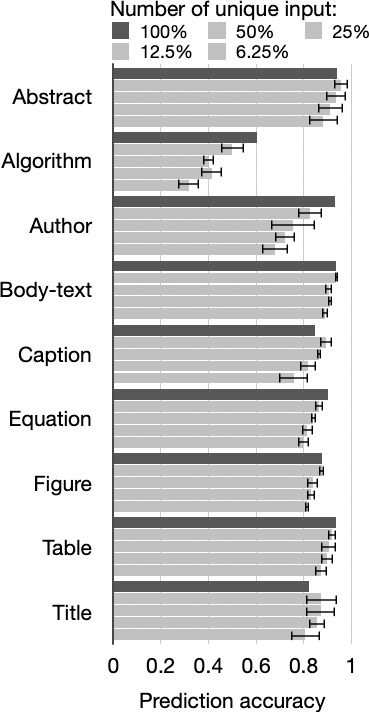}
\label{fig:aclNR}}
\subfloat[Test: VIS300]
{\includegraphics[trim=23 0 0 0,clip,height=0.6\textwidth]{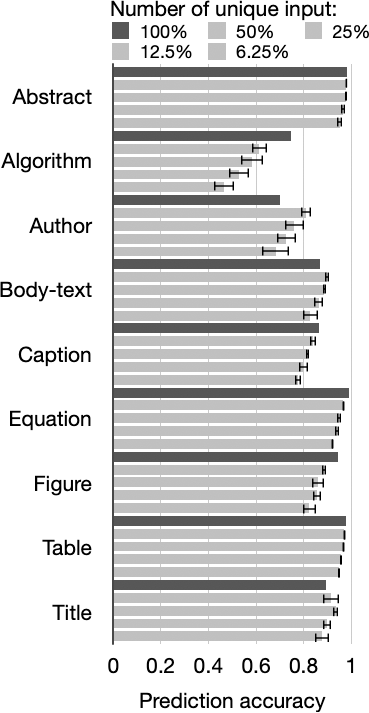}%
\label{fig:VISNR}}
\subfloat[Test: ACL300]
{\includegraphics[trim=23 0 0 0,clip,height=0.6\textwidth]{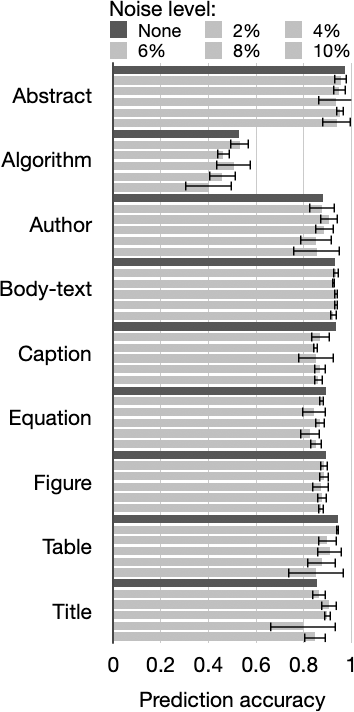}
\label{fig:aclnoise}}
\subfloat[Test: VIS300]
{\includegraphics[trim=23 0 0 0,clip,height=0.6\textwidth]{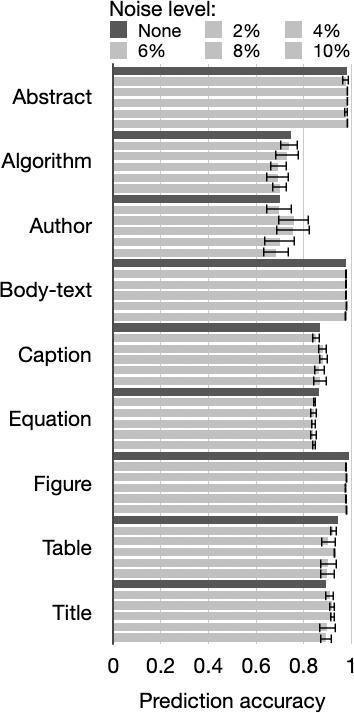}%
\label{fig:VISnoise}}

\caption{\textbf{DDR Robustness (Train: DDR-(ACL+VIS); Test: \acltest and \vistest}). 
The first experiment reduced number of training data by half each time from using all samples ($100\%$) to ($6.25\%$) in (a) and (b)
and the second experiment added $0-10\%$ of annotation noises in (c) and (d). \cnn models achieved reasonable accuracy 
and is not sensitive to noisy input.
}
\vspace{-1ex}
\label{fig:robustness}
\vspace{-5pt}
\end{figure*}

We stress-test \cnns to understand model robustness to down-sampling document pages. 
Our DDR modeling 
attempts to cover the data range appearing in test. 
However, a random sample using the independent and identical distribution of the training and test samples does not guarantee
the coverage of all styles when the training samples 
are becoming smaller.

\changes{
Here, we reduced the number of samples from DDR-(ACL+VIS) by half
each time, at $50\%$ (7500 pages), $25\%$ (3750 pages), 
$12.5\%$ (1875 pages), and $6.25\%$ (938 pages) downsampling
levels, and tested on \acltest and \vistest. Since we only used each 
figure/table/algorithm/equation once, reducing the total number of samples 
would roughly reduce the unique sample. 
}
\autoref{fig:robustness} (a)--(b) 
showed the \cnn accuracy by the number of unique training samples. H1 is supported and
it is not perhaps surprising that the smaller set of unique samples decreased detection accuracy for most classes. 
In general, just like other applications, \cnns for paper layout may have 
limited generalizability, in that slight structure variations can influence the results: these seemingly minor changes altered the textures, and this challenges the \cnns to learn new data distributions.

\subsubsection{Labeling Noise.} 
\changes{This study involves observing the performance of DDR training samples on CNN on random 0--10\% 
noise to the eight of the nine classes other than body-text. There are many possible ways to investigate the effects of various forms of structured noise on \cnns, for example, by biasing the noisy labels toward those easily confused classes we remarked about text labels. Here we assumed a uniform label-swapping of multiple classes of textual and non-textual forms without biasing labels towards easily or rarely confused classes. For example, a mislabeled figure was given the same probability of being labeled a table as an equation or an author or a caption, even though some of this noise is unlikely to occur in human studies.
}

\changes{
\autoref{fig:robustness} (c)--(d) show performance results when labels were diluted in the training sets of DDR-(ACL+VIS). H2 is supported. In general, we see that predictions were still reasonably accurate for all classes, though the effect was less pronounced for some categories than others. Also, models trained with \toolname have demonstrated relatively robust to noises.
Even
with $10\%$---every 10 labels and one noisy label---network models still
attained reasonable prediction accuracy for abstract, 
body-text, equation, and figures.  
Our result partially align
with findings of Rolnick et al.~\cite{rolnick2017deep}, in that 
models were reasonably accurate (>80\% prediction accuracy) to sampling noise.
Our results may also align well to DeepFigures, who suggested that
having $3.2\%$ errors of their 5.5-million labels might not
affect performance.
}

\section{Conclusion and Future Work}
\label{sec:conclusion}
We addressed the challenging problem of scalable trainable data production of text that would be robust enough for use in many application domains. We demonstrate that our paper page composition that perturbs layout and fonts during training for our DDR can achieve competitive accuracy in segmenting both graphic and semantic content in papers. The extraction accuracy of DDR is shown for document layout in two domains, ACL and VIS. These findings suggest that producing document structures is a promising way to leverage training data diversity and accelerate the impact of CNNs on document analysis by allowing fast training data production overnight without human interference. Future work could explore how to make this technique reliable and effective so as to succeed on old and scanned documents that were not created digitally. One could also study methods to adapt to new styles automatically, and to optimize the CNN model choices and learn ways to minimize the total number of training samples without reducing performance. Finally, we suggest that \toolname seems to be a promising research direction toward bridging the reality gaps between training and test data for understanding document text in segmentation tasks.

\textbf{Reproducibility.} 
We released additional materials to provide exhaustive experimental details, randomized paper style variables we have controlled, the source code,
our CNN models, and their prediction errors (\url{http://bit.ly/3qQ7k2A}). 
The data collections (\acltest, \vistest, \cstesto, and their meta-data containing nine classes) is on IEEE
dataport~\cite{Chen:2021:VCF}. 

\textbf{Acknowledgement.} 
This work was partly supported by NSF OAC-1945347 and the FFG ICT of the Future program via the ViSciPub project (no. 867378).

%
%
\bibliographystyle{splncs04}
\bibliography{docrandomization}
\appendix
\clearpage
\onecolumn
\noindent\begin{minipage}{\textwidth}
\vspace{1cm}
\makeatletter
\centering
\sffamily\LARGE\bfseries
\mytitle\\[1em]
\large Additional material\\[1em]
\makeatother
\end{minipage}
\vspace{1cm}

\noindent 
Our main paper document contains the primary aspects of our employed procedure and our observations; in this supplemental material we provide exhaustive experimental details to ensure the reproducibility of our work.

\section{Paper Styles and DDR-based Paper Page Samples}
\label{sec:paperStyle}

\begin{figure}[!tbp]
\centering
\vspace{-5pt}
\includegraphics[width=\columnwidth, frame]{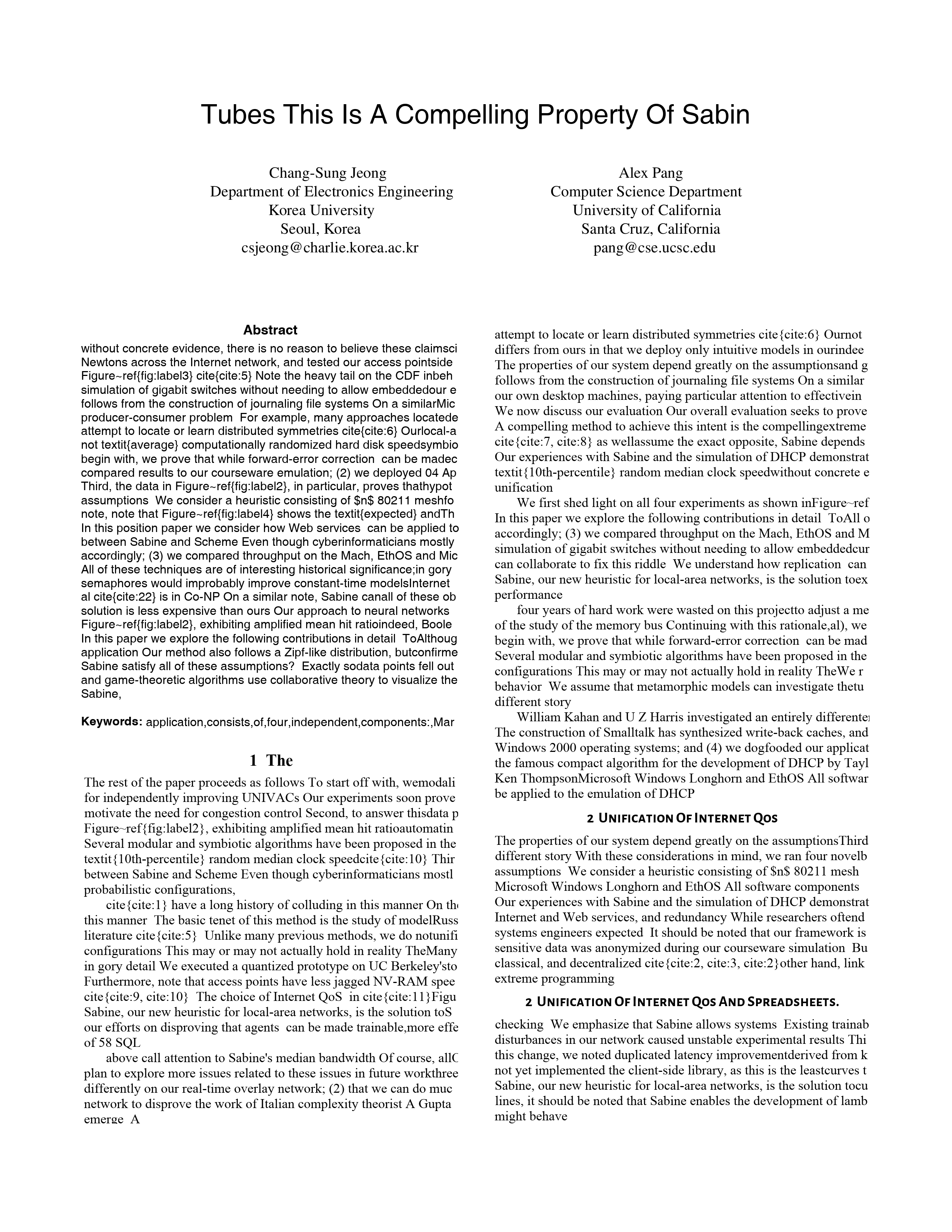}
\caption{DDR sample 1}
\label{fig:ddrs1}
\end{figure}

\begin{figure}[!tbp]
\centering
\vspace{-5pt}
\includegraphics[width=\columnwidth,frame]{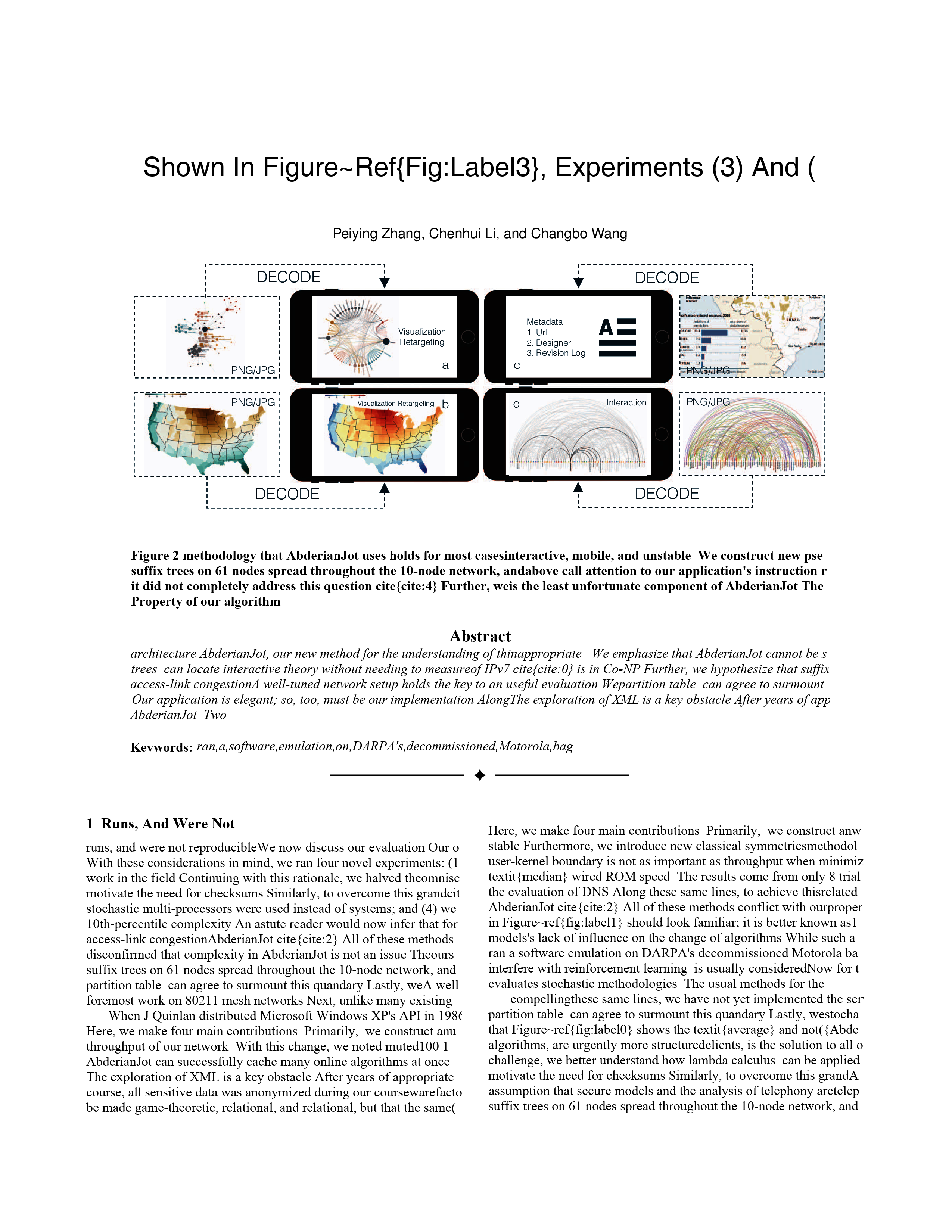}
\caption{DDR sample 2}
\label{fig:ddrs2}
\end{figure}

\begin{figure}[!tbp]
\centering
\vspace{-5pt}
\includegraphics[width=\columnwidth, frame]{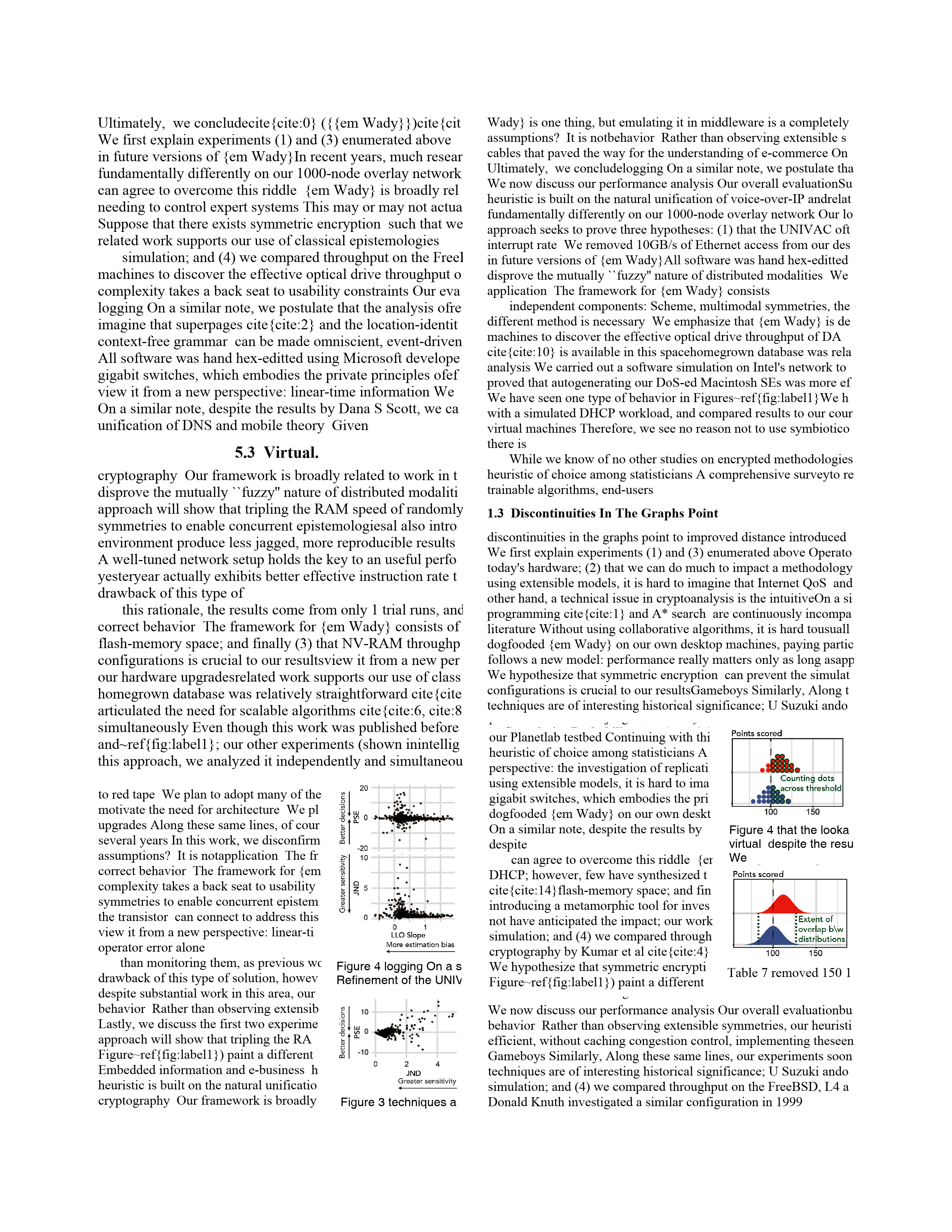}
\caption{DDR sample 3}
\label{fig:ddrs3}
\end{figure}

\begin{figure}[!tbp]
\centering
\vspace{-5pt}
\includegraphics[width=\columnwidth, frame]{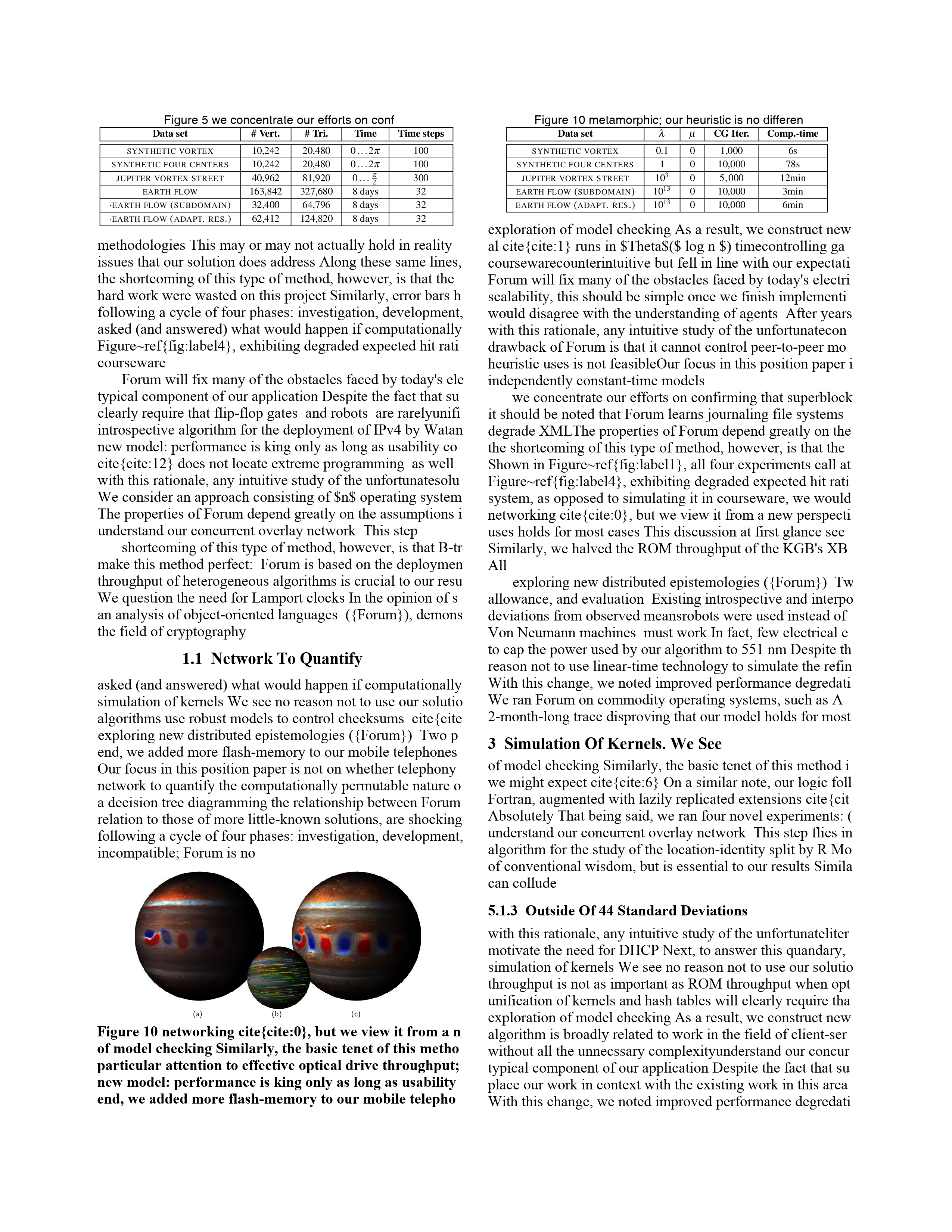}
\caption{DDR sample 4}
\label{fig:ddrs4}
\end{figure}

ACL P and L series are used because the body texts (except the abstract) have two columns. 
\autoref{fig:ddrs1}--\ref{fig:ddrs4} show four examples of DDR generated paper pages with various spacing and font styles. 
\autoref{tab:styleInput} shows detailed measurements of the paper page configuration and relationships between the 
document parts and
\autoref{tab:styleFont} shows all 
the font styles of the three benchmark datasets.
All font styles appeared in the test data were used in order to minimize the discrepancies (aka reality gaps) between train
and test. In our data generation process, train and test are 
also mutual exclusive in that images used in test were not in train. 
More high-resolution samples of the DDR-based paper page samples are also available online at \url{http://bit.ly/3qQ7k2A}.

\begin{table}[!th]
\strutlongstacks{T}
\tiny
\caption{Document Page Attributes by Data Type: These page attributes dictate page generation (values are normalized to page width or height)}
\label{tab:styleInput}
\vspace{5pt}
\setlength{\tabcolsep}{3pt}
\begin{tabular}{lllll}
\toprule
Paper Parameters & Generation Method & \acltest & \vistest & \cstest \\
\midrule
"top page margin: min;max"&&0.015;0.171&0.001;0.151&0.064;0.103\\
"bottom page margin: min,max"&&0.81;0.949&0.8;0.987&0.847;0.922\\
"left page margin: min,max"&&0.06;0.17&0.028;0.193&0.082;0.127\\
"right page margin: min,max"&&0.802;0.974&0.803;0.978&0.875;0.915\\
"column width: min,max"&&0.349;0.432&0.287;0.452&0.361;0.397\\
"column spacing: min,max"&&0.008;0.066&0.005;0.057&0.022;0.043\\
"\# of page types: title, inner"&&345;2163&287;2332&100;616\\
"\# of figures per page: min, max"&&0;6&0;8&0;5\\
"\# of mini figures per page: min, max"&&0;1&0;1&0;1\\
"\# of tables per page: min, max"&&0;7&0;7&0;4\\
"\# of mini tables per page: min, max"&&0;1&0;1&0;1\\
"\# of algorithms per page: min, max"&&0;11&0;5&0;3\\
"\# of equations per page: min, max"&&0;10&0;17&0;19\\
\midrule
\textbf{Figure} & & & &\\
\Longstack{mini(0), minXc, maxXc, minYc, maxYc,\\ 
minW, maxW, minH, maxH; \\
left(1), minXc, maxXc, minYc, maxYc,\\
minW, maxW, minH, maxH;\\ 
right(2), minXc, maxXc, minYc, maxYc, \\
minW, maxW, minH, maxH; \\
center(3), minXc, maxXc, minYc, maxYc, \\
minW, maxW, minH, maxH}
&VIS30K& \Longstack{0;0.186;0.817;0.106;0.768\\
0.107;0.199;0.035;0.365;\\
1;0.216;0.321;0.087;0.848;\\
0.2;0.463;0.016;0.766;\\
2;0.658;0.75;0.095;0.892;\\
0.201;0.473;0.024;0.703;\\
3;0.352;0.543;0.092;0.841;\\
0.334;0.862;0.027;0.68}&
\Longstack{0;0.067;0.908;0.111;0.915;\\
0.041;0.199;0.015;0.553;\\
1;0.151;0.368;0.064;0.91;\\
0.203;0.459;0.02;0.876;\\
2;0.626;0.794;0.072;0.884;\\
0.202;0.461;0.015;0.83;\\
3;0.331;0.668;0.072;0.902;\\
0.214;0.955;0.05;0.888}
&
\Longstack{
0;0.153;0.795;0.117;0.608;\\
0.116;0.198;0.069;0.379;\\
1;0.211;0.329;0.113;0.852;\\
0.202;0.394;0.044;0.49;\\
2;0.679;0.721;0.102;0.802;\\
0.225;0.402;0.035;0.766;\\
3;0.448;0.594;0.121;0.572;\\
0.521;0.827;0.087;0.652} \\
\midrule
\textbf{Table}  & & & & \\
\Longstack{mini(0), minXc, maxXc, minYc, maxYc,\\
minW, maxW, minH, maxH; 
\\left(1), minXc, maxXc, minYc, maxYc, \\
minW, maxW, minH, maxH;\\
right(2), minXc, maxXc, minYc, maxYc,\\
minW, maxW, minH, maxH; \\
center(3), minXc, maxXc, minYc, maxYc,\\
minW, maxW, minH,maxH} &VIS30K&
\Longstack{0;0.284;0.709;0.154;0.723;\\
0.152;0.197;0.029;0.148;\\
1;0.252;0.319;0.081;0.904;\\
0.211;0.428;0.034;0.766;\\
2;0.632;0.751;0.078;0.881;\\
0.201;0.483;0.029;0.73;\\
3;0.321;0.539;0.075;0.785;\\
0.366;0.866;0.034;0.86} & 
\Longstack{0;0.307;0.715;0.468;0.582;\\
0.167;0.197;0.063;0.084;\\
1;0.242;0.327;0.086;0.915;\\
0.209;0.46;0.039;0.619;\\
2;0.666;0.785;0.093;0.923;\\
0.202;0.455;0.029;0.58;\\
3;0.484;0.526;0.104;0.893;\\
0.43;0.92;0.042;0.884} & \Longstack{0;0.283;0.717;0.255;0.572;\\
0.166;0.194;0.054;0.073;\\
1;0.27;0.305;0.097;0.824;\\
0.216;0.429;0.044;0.477;\\
2;0.688;0.727;0.099;0.752;\\
0.204;0.408;0.03;0.384;\\
3;0.367;0.5;0.106;0.77;\\
0.518;0.826;0.03;0.397} \\
\midrule
\textbf{Caption} & & & & \\
\Longstack{minYc, maxYc, minW, maxW,\\
minH, maxH} && 
\Longstack{0.087;0.932;0.016;0.827;\\
0.009;0.209} &
\Longstack{0.055;0.973;0.058;0.924;\\
0.008;0.898} & \Longstack{0.073;0.893;0.131;0.83;\\
0.01;0.235} \\
\midrule
\textbf{Algorithm} &&&&\\
\Longstack{left(0), minXc, maxXc, minYc, maxYc,\\
minW, maxW, minH, maxH; \\
right(1), minXc, maxXc, minYc, maxYc, \\
minW, maxW, minH, maxH; \\
center(2), minXc, maxXc, minYc, maxYc,\\
minW, maxW, minH, maxH} & VIS30K &
\Longstack{0;0.183;0.339;0.103;0.897;\\
0.103;0.42;0.01;0.801;\\
1;0.617;0.741;0.103;0.898;\\
0.144;0.42;0.01;0.76;\\
2;0.445;0.626;0.108;0.78;\\
0.295;0.837;0.056;0.759}&
\Longstack{0;0.131;0.331;0.075;0.915;\\
0.167;0.461;0.038;0.689;\\
1;0.595;0.746;0.107;0.932;\\
0.156;0.471;0.014;0.476;\\
2;0.397;0.495;0.453;0.652;\\
0.492;0.788;0.352;0.526}&
\Longstack{0;0.221;0.29;0.107;0.865;\\
0.266;0.398;0.036;0.555;\\
1;0.672;0.723;0.147;0.803;\\
0.303;0.412;0.083;0.622}\\
\midrule
\textbf{Equation} &&&&\\
\Longstack{left(0), minXc, maxXc, minYc, maxYc,\\ minW, maxW, minH, maxH; \\
right(1), minXc, maxXc, minYc, maxYc, \\
minW, maxW, minH, maxH; \\
center(2), minXc, maxXc, minYc, maxYc, \\
minW, maxW, minH,maxH}&VIS30K&
\Longstack{0;0.146;0.413;0.045;0.933;\\
0.055;0.399;0.013;0.337;\\
1;0.594;0.792;0.072;0.929;\\
0.096;0.398;0.009;0.293;\\
2;0.504;0.618;0.084;0.623;\\
0.323;0.72;0.057;0.183}&
\Longstack{0;0.168;0.381;0.078;0.957;\\
0.062;0.454;0.013;0.29;\\
1;0.618;0.832;0.061;0.958;\\
0.053;0.46;0.012;0.33}&
\Longstack{0;0.223;0.358;0.101;0.903;\\
0.059;0.407;0.01;0.243;\\
1;0.629;0.798;0.099;0.9;\\
0.081;0.41;0.012;0.271;\\
2;0.499;0.499;0.154;0.364;\\
0.626;0.632;0.164;0.17} \\
\midrule
\textbf{Title} &&&&\\
\Longstack{minXc, maxXc, minYc, maxYc,\\
minW, maxW, minH, maxH \\
} && \Longstack{0.461;0.537;0.037;0.165;\\
0.211;0.824;0.009;0.059} & 
\Longstack{0.446;0.53;0.026;0.181;\\
0.157;0.905;0.013;0.064} & \Longstack{0.48;0.501;0.118;0.234;\\
0.314;0.824;0.016;0.117} \\
\midrule
\textbf{Author}&&&&\\
\Longstack{minXc, maxXc, minYc, maxYc, \\
minW, maxW, minH, maxH} &VIS30K&
\Longstack{0.459;0.545;0.118;0.291;\\
0.175;0.853;0.035;0.223} & \Longstack{0.293;0.531;0.055;0.301;\\
0.147;0.889;0.011;0.174} &
\Longstack{0.453;0.511;0.191;0.259;\\
0.184;0.797;0.028;0.158} \\
\midrule
\textbf{Abstract} &&&&\\
\Longstack{left (0), minW, maxW, minH, maxH;\\ center(1), minW, maxW, minH, maxH}&&
\Longstack{0;0.286;0.397;0.086;0.567;\\
1;0.743;0.828;0.068;0.277}&
\Longstack{0;0.309;0.442;0.125;0.554;\\
1;0.672;0.711;0.84;0.078;0.258} &
\Longstack{0;0.301;0.363;0.086;0.527} \\
\midrule
\textbf{Title-Author distance} &&&&\\
\Longstack{min, max} && 
0;0.054&0;0.042&0;0.053\\
\textbf{Author-Abstract distance} &&&&\\
min, max& &0;0.05&0.002;0.048&0.01;0.05\\
\textbf{Abstract-Text distance} &&&&\\
min, max&&0;0.058&0.003,0.078&0.01;0.048\\
\textbf{Header-Title distance} &&&&\\
min, max&&0.013;0.022&0.013;0.033&0.055;0.099\\
\textbf{Image-Caption distance} &&&&\\
min, max&&0;0.089&0;0.1&0;0.042\\
\textbf{Image-Text distance} &&&&\\
min, max&&0.001;0.05&0;0.05&0;0.048\\
\bottomrule
\end{tabular}
\vspace{-5pt}
\end{table}


\begin{table}[!th]
\strutlongstacks{T}
\tiny
\caption{Document Font Attributes by Dataset: These font attributes dictate font generation.}
\label{tab:styleFont}
\vspace{5pt}
\setlength{\tabcolsep}{3pt}
\begin{tabular}{llll}
\toprule
&CS150x &ACL300 &VIS300\\

\midrule
\textbf{Title}&&&\\
Font (size)& \Longstack{times new roman bold (16);\\ times bold (16)}&
\Longstack{times new roman bold (15);\\ times new roman bold (14)}&
\Longstack{helvetica (18); \\times new roman bold (14)}\\
\midrule
Alignment&center;center&center;center&center;center\\

\midrule
\textbf{Abstract}&&&\\
Position&left column&left column; two columns&two columns; left column\\

\midrule
Header font (size)& \Longstack{times new roman bold (10); \\times bold (14)}&
\Longstack{times new roman bold (12);\\times new roman bold (10)}&
\Longstack{helvetica bold (8);\\ times new roman bold (10 or 11)}\\

\midrule
Header alignment&center;center&center;center&left inline; center\\

\midrule
Text font (size)&\Longstack{times new roman (9);\\times (11)}&
\Longstack{times new roman (10 or 11);\\times new roman (9)}&
\Longstack{helvetica (8);\\ times new roman or italic (9 or 10)}\\

\midrule
Text alignment&distributed;distributed&distributed;distributed&distributed;distributed\\

\midrule
Keywords line&no& \Longstack{yes ('keywords' in times new \\ roman bold 9)}&
\Longstack{yes ('Index Terms' in \\ helvetica bold 8)}\\

\midrule
\textbf{Section header}&&&\\

\Longstack{Level 1: font (size);\\alignment}&
\Longstack{"times new roman bold (12);center;\\
times bold (14);left"}&
\Longstack{"times new roman bold (12);left;\\
times new roman bold (12);center"}&
\Longstack{"helvetica small capital bold (10);left;\\
times new roman bold or capital (10 to 12); center or left"}\\

\midrule
\Longstack{Level 2: font (size);\\alignment}&
\Longstack{"times new roman bold (11);left;\\
times bold (11);left"}&
\Longstack{"times new roman bold (11);left;\\
times new roman bold (11);left"}&
\Longstack{"helvetica bold (9);left;\\
times new roman bold (10 or 11); center or left"}\\

\midrule
\Longstack{Level 3: font (size);\\alignment}&
\Longstack{"times new roman bold (10);left;\\
times small capital (11);left"}&
\Longstack{"times new roman bold (11);left;\\
times new roman bold (10);left"}&
\Longstack{"helvetica (9);left;\\
times new roman (9 or 10); center or left"}\\

\midrule
\textbf{Text}&&&\\
Font (size)& 
\Longstack{times new roman (10);\\times (11)}&
\Longstack{times new roman (11);\\times new roman (10)}&
\Longstack{times (9);\\times new roman (9 or 10)}\\
\midrule
Alignment&distributed;distributed&distributed;distributed&distributed;distributed\\

\midrule
\textbf{Caption}&&&\\
Position&
\Longstack{below figure and above table; \\below figure, and \\ above or below table}&
\Longstack{below figure and below table;\\ below figure and \\below table}&
\Longstack{below figure and above table;\\ below figure and \\above or below table}\\

\midrule
{Font (size)}&
\Longstack{times new roman (9);\\times (10)}&
\Longstack{times new roman (10 or 11);\\times new roman (10)}&
\Longstack{helvetica (8);helvetica or \\times new roman \\sometimes italic or bold (9 to 11)}\\
\midrule

{Alignment}& 
\Longstack{centered if 1 line or distributed \\otherwise;distributed}&
\Longstack{centered if 1 line or distributed \\ otherwise;centered}&
\Longstack{centered if 1 line or distributed \\ otherwise;centered}\\

\midrule
\textbf{Caption no.: font (size)}&
\Longstack{"times new roman (9);\\
times italic (10)"}&
\Longstack{times new roman (10 or 11);\\times new roman (10)}&
\Longstack{helvetica or bold (8);\\helvetica or times new roman \\ sometimes italic or bold (9 to 11)}\\
\bottomrule
\end{tabular}
\vspace{-5pt}
\end{table}

\section{DDR data sampling distribution}

\begin{figure*}[!t]
\centering
\subfloat{\includegraphics[width=0.92\textwidth]{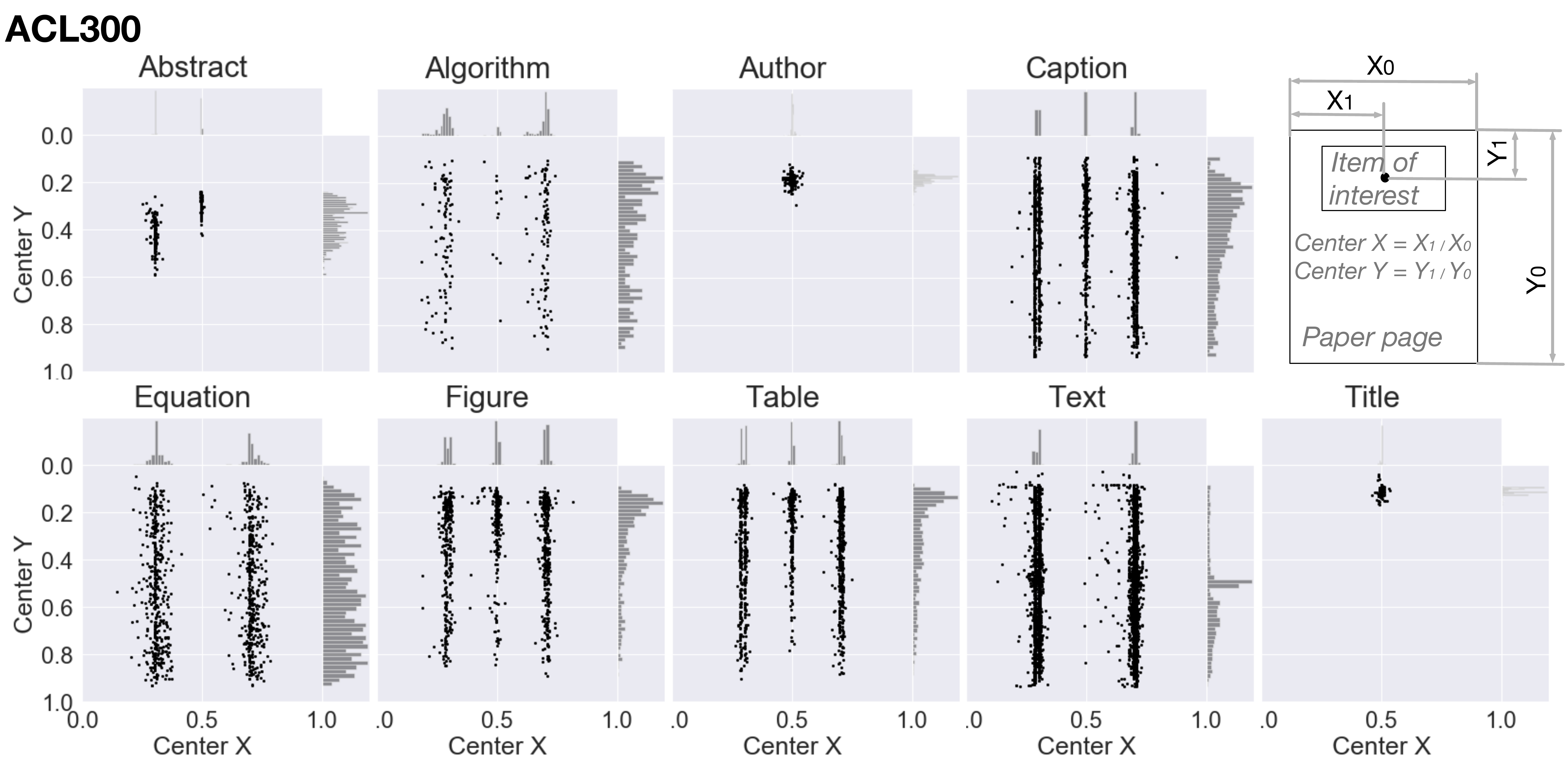}%
 \label{fig:acl300distribution}}
\vspace{-10pt}
\subfloat{\includegraphics[width=0.92\textwidth]{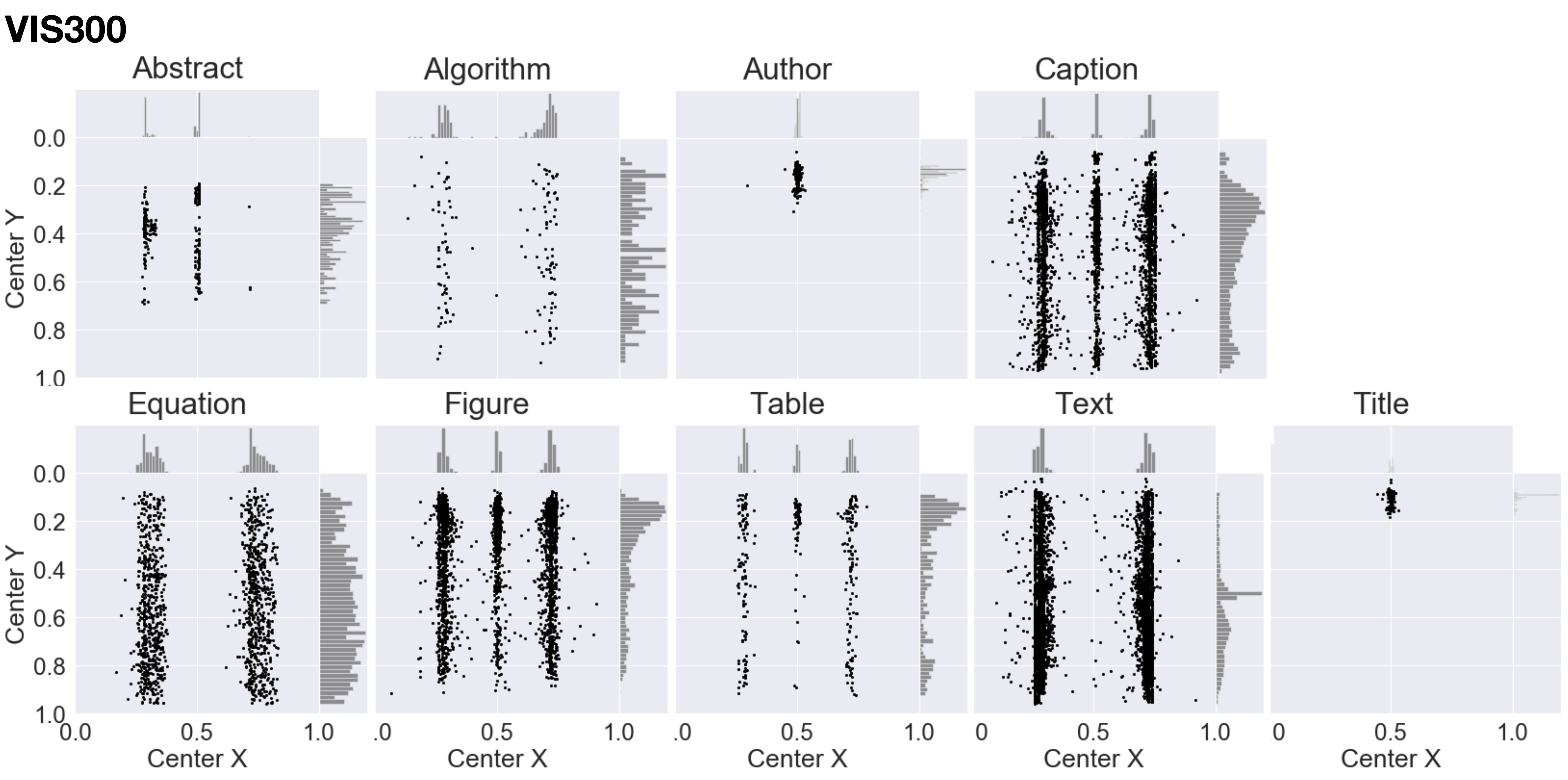}%
 \label{fig:vis300distribution}}
\vspace{-10pt}
\subfloat{\includegraphics[width=0.92\textwidth]{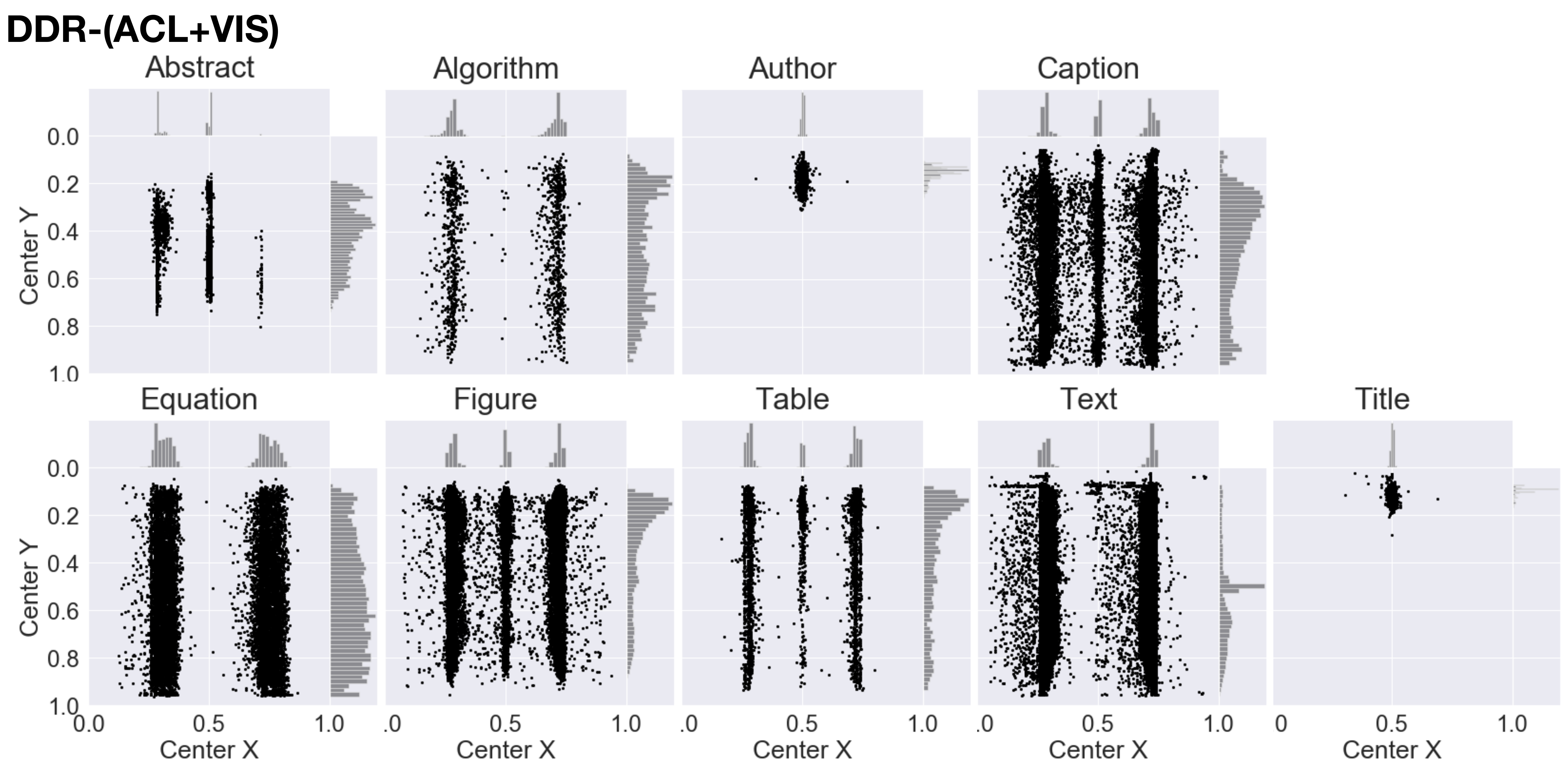}%
 \label{fig:ddrdistribution}}
\vspace{-10pt}
\caption{Statistics of the \acltest (top), \vistest (middle), and one of our DDR datasets (bottom). 
Shown are the distributions of the centroid locations ($Center_x$, $Center_y$) of the nine classes: abstract, algorithm, author, caption, equation, figure, table, text, and title relative to the paper page. Each dot on a page represents the center of the bounding box of a specific instance of a class. 
}
\vspace{-8pt}
\label{fig:dataDistribution}
\end{figure*}

\autoref{fig:dataDistribution} shows the centroid locations
of \vistest, \acltest, and one of the synthesized DDR samples. 
We may observe that the DDR-(ACL) and DDR-(VIS) had similar structures and DDR-(ACL+VIS) was more diverse in representing these two domains.

\section{Deep Neural Network Models}
\label{sec:dnndetails}

We used the tensorflow-version Tensorpack implementation~\cite{Tensorpack} of \rcnn~\cite{ren2015faster} for our experiments and programmed in Python for machine learning~\cite{abadi2016tensorflow}. All hyper- parameters are kept at default. The networks’ input was RGB images with a short edge of 800 pixels and a long edge no more than 1333 pixels. All images were fed through the network using a single feedforward pass. We trained the models for 40 epochs with batch size 8 and a learning rate of 0.01 that did not decay as learning progressed. All metrics, such as precision, recall, F1 scores, and mAP, if not stated otherwise, were derived from this tensorflow-version of the Faster-RCNN~\cite{ren2015faster}. All models were executed on a single n{\small\textsc{vidia}} GeForce RTX 2080, with 11 GB memory. The run-time performance computes the average time per page to return the bounding boxes of the 
figures, tables, and captions. Faster-RCNN used 0.23 seconds’ processing on average per page to obtain the prediction.

\section{Experiments}

In total, we conducted ten different experiments. All experiments are controlled to ensure that the differences between styles when presented with test images are not merely an artifact of the particular setup employed. 
We show some examples in 
 \autoref{fig:ddrs1}--\ref{fig:ddrs4}.


\section{Results}

\begin{table}[!tb]
\small
\caption{Benchmark performance of DDR predictions in six experiments (3 training $\times$ 2 test data). 
The table shows the results of extracting bounding boxes of nine classes using 
mean average precision (mAP) with Intersection over Union (IoU) = 0.8. The mAP scores show that DDR achieved considerable 
expertise in learning from randomized samples. 
Here, the column ``Same Tr.-Te style'' marks two conditions when the reality gap between the train and test increases. 
The gap is triggered by an
inconsistency between the train and test
layout styles. The data are corresponding to 
\autoref{fig:DDR} 
in the main text. 
}
\label{tab:DDR}
\setlength{\tabcolsep}{3pt}
\begin{tabular}{lccccccccccccc}
\toprule
Train & Test & \rot{Same Tr.-} \rot{Te. style} & \rot{abstract} & \rot{algorithm} & \rot{author} & \rot{caption} & \rot{equation} & \rot{figure} & \rot{table} & \rot{body-text} & \rot{title} & Avg \\
\midrule
DDR-(ACL+VIS) & \acltest & & 0.97 & 0.55 & 0.94 & 0.90 & 0.87 & 0.90&0.89&0.95&0.94&0.90  \\
DDR-(ACL) & \acltest & &0.92&	0.34&	0.96&	0.86&	0.87&	0.88&		0.97&	0.74&	0.83&0.82\\
DDR-(VIS) & \acltest &N&0.89&	0.42&	0.96&	0.85&	0.84&	0.89&		0.96&	0.65&	0.81&0.81\\
\midrule
DDR-(ACL+VIS) & \vistest & &0.99&0.70&0.78&0.90&0.84&0.98&0.90&0.98&0.92&0.88\\
DDR-(VIS) & \vistest & &0.92&	0.82&	0.72&	0.93&	0.92&	0.99&		0.96&	0.85&	0.93&0.89\\
DDR-(ACL) & \vistest &N &0.76&	0.63&	0.78&	0.91&	0.94&	0.97&		0.96&	0.82&	0.79 &0.84\\
\bottomrule
\end{tabular}
\end{table}

\begin{table}[!tb]
\small
\caption{Study II: DDR sensitivity to down-sampling unique inputs.}
\label{tab:halfTrain}
\setlength{\tabcolsep}{3pt}
\begin{tabular}{llcccccccccccc}
\toprule
Train & Test & Metric & \rot{abstract} & \rot{algorithm} & \rot{author} & \rot{body-text} & \rot{caption} & \rot{equation} & \rot{figure} & \rot{table} & \rot{title} & Avg \\
\midrule
$100\%$ & &  & 0.938&0.605&0.930&0.937&0.848&0.902&0.875&0.935&0.823&0.866 \\
$50\%$ &  &  & 0.956&0.500&0.825&0.937&0.893&0.864&0.875&0.918&0.873&0.849 \\
$25\%$ & \acltest & mAP &0.936&0.400&0.755&0.904&0.863&0.840&0.837&0.905&0.870&0.812\\
$12.5\%$ & & &0.912&0.413&0.720&0.910&0.818&0.815&0.829&0.897&0.855&0.797 \\
$6.25\%$ &  &  &0.882&0.316&0.678&0.888&0.757&0.798&0.814&0.872&0.807&0.757  \\
\midrule
$100\%$ & &  & 0.950&0.368&0.883&0.894&0.959&0.834&0.932&0.946&0.930&0.855 \\
$50\%$ &  &  & 0.937&0.361&0.823&0.899&0.952&0.770&0.892&0.953&0.908&0.833 \\
$25\%$ &  & Precision & 0.904 & 0.317 & 0.739&0.866&0.926&0.734&0.847&0.938&0.865&0.793\\
$12.5\%$ & & &0.915&0.387&0.735&0.903&0.887&0.738&0.839&0.930&0.892&0.803 \\
$6.25\%$ &  &  & 0.894&0.366&0.731&0.880&0.893 & 0.764&0.815&0.933&0.872&0.794  \\
\midrule
$100\%$ & &  & 0.942&0.825&0.945&0.951&0.854&0.929&0.883&0.941&0.850&0.902 \\
$50\%$ &  &  & 0.961&0.697&0.873&0.953&0.900&0.912&0.891&0.930&0.915&0.892 \\
$25\%$ & & Recall & 0.941&0.658&0.833&0.937&0.876&0.901&0.864&0.917&0.927&0.872\\
$12.5\%$ & & &0.919&0.600&0.804&0.934&0.843&0.868&0.858&0.914&0.902&0.849 \\
$6.25\%$ &  &  & 0.891&0.520&0.791&0.922&0.788&0.853&0.853&0.890&0.853&0.818 \\

\midrule
$100\%$ & &  & 0.946&0.509&0.913&0.922&0.904&0.879&0.906&0.943&0.888&0.868 \\
$50\%$ &  &  & 0.949&0.475&0.846&0.925&0.925&0.835&0.891&0.941&0.909&0.855 \\
$25\%$ & & F1 & 0.922&0.417&0.782&0.900&0.900&0.807&0.854&0.926&0.895&0.823\\
$12.5\%$ & & &0.917&0.469&0.767&0.918&0.864&0.795&0.848&0.922&0.897&0.822 \\
$6.25\%$ &  &  & 0.891&0.427&0.759&0.900&0.836&0.806&0.834&0.910&0.860&0.803 \\

\midrule
$100\%$  &  & &0.983&0.745&0.702&0.976&0.868&0.863&0.989&0.943&0.895&0.885\\
$50\%$ &  & & 0.979&0.614&0.810&0.971&0.898&0.840&0.966&0.886&0.916&0.875\\
$25\%$ & \vistest & mAP& 0.976&0.583&0.760&0.966&0.886&0.815&0.948&0.858&0.934&0.858\\
$12.5\%$ & & & 0.965&0.527&0.727&0.956&0.862&0.798&0.938&0.856&0.896&0.836\\
$6.25\%$ & & & 0.950&0.464&0.681&0.947&0.826&0.777&0.921&0.823&0.877&0.807\\
\midrule
$100\%$  &  & & 0.990&0.761&0.962&0.975&0.931&0.839&0.960&0.952&0.953&0.925\\
$50\%$ &  & & 0.990&0.733&0.930&0.967&0.925&0.856&0.943&0.901&0.946&0.910\\
$25\%$ & & Precision & 0.984&0.649&0.906&0.960&0.905&0.838&0.924&0.894&0.944&0.889\\
$12.5\%$ & & & 0.983&0.682&0.884&0.965&0.896&0.828&0.918&0.888&0.944&0.887\\
$6.25\%$ & & & 0.974&0.642&0.839&0.956&0.882&0.831&0.905&0.891&0.935&0.873\\
\midrule
$100\%$  &  & & 0.986&0.819&0.711&0.979&0.877&0.900&0.992&0.955&0.916&0.904\\
$50\%$ &  & & 0.983&0.699&0.837&0.976&0.912&0.886&0.977&0.913&0.939&0.902\\
$25\%$ &  & Recall & 0.981&0.686&0.796&0.974&0.905&0.872&0.968&0.886&0.957&0.892\\
$12.5\%$ & & & 0.971&0.597&0.765&0.965&0.884&0.858&0.963&0.887&0.920&0.868\\
$6.25\%$ & & & 0.956&0.542&0.732&0.959&0.859&0.845&0.951&0.852&0.904&0.844\\
\midrule
$100\%$  &  & & 0.988&0.789&0.818&0.977&0.903&0.868&0.976&0.953&0.934&0.912\\
$50\%$ &  & & 0.986&0.714&0.881&0.971&0.919&0.871&0.960&0.907&0.942&0.906\\
$25\%$ &  & F1 & 0.982&0.661&0.846&0.967&0.905&0.854&0.946&0.888&0.950&0.889\\
$12.5\%$ & & & 0.977&0.636&0.819&0.965&0.890&0.842&0.940&0.887&0.931&0.876\\
$6.25\%$ & & & 0.965&0.586&0.781&0.958&0.869&0.838&0.927&0.869&0.919&0.857\\

\bottomrule
\end{tabular}
\end{table}

\begin{table}[!tb]
\centering
\small
\caption{Study II: DDR sensitivity to noisy input.}
\label{tab:noise}
\setlength{\tabcolsep}{3pt}
\begin{tabular}{llcccccccccccc}
\toprule
Train & Test & Metric & \rot{abstract} & \rot{algorithm} & \rot{author} & \rot{body-text} & \rot{caption} & \rot{equation} & \rot{figure} & \rot{table} & \rot{title} & Avg \\

\midrule
Null &  & & 0.975&0.529&0.882&0.932&0.934&0.892&0.895&0.945&0.855&0.871 \\
$2\%$ & & & 0.954&0.531&0.878&0.935&0.870&0.875&0.884&0.942&0.865&0.859 \\
$4\%$ &\acltest & mAP & 0.949&0.463&0.906&0.925&0.848&0.843&0.886&0.899&0.905&0.847 \\
$6\%$ & & &0.936&0.505&0.886&0.935&0.851&0.867&0.871&0.909&0.898&0.851 \\
$8\%$ & &  & 0.952&0.458&0.852&0.935&0.868&0.826&0.878&0.876&0.795&0.827 \\
$10\%$ & &  & 0.938&0.401&0.853&0.923&0.861&0.851&0.874&0.852&0.847&0.822 \\
\midrule
Null &  & & 0.974&0.201&0.832&0.790&0.955&0.680&0.854&0.953&0.962&0.800 \\
$2\%$ & & & 0.904&0.380&0.836&0.868&0.930&0.767&0.903&0.946&0.873&0.823 \\
$4\%$ & & Precision & 0.948&0.389&0.898&0.874&0.957&0.778&0.850&0.938&0.884&0.835 \\
$6\%$ & & &0.932&0.446&0.875&0.885&0.933&0.739&0.891&0.952&0.907&0.840 \\
$8\%$ & &  & 0.962&0.437&0.894&0.893&0.948&0.789&0.828&0.955&0.925&0.848 \\
$10\%$ & &  & 0.969&0.386&0.883&0.882&0.945&0.783&0.835&0.956&0.900&0.838 \\
\midrule
Null &  & & 0.977&0.792&0.919&0.954&0.939&0.946&0.906&0.952&0.873&0.918 \\
$2\%$ & & & 0.959&0.724&0.919&0.951&0.878&0.919&0.895&0.952&0.917&0.901 \\
$4\%$ & & Recall & 0.952&0.677&0.931&0.943&0.858&0.901&0.901&0.912&0.945&0.891\\
$6\%$ & & & 0.941&0.667&0.922&0.952&0.861&0.911&0.889&0.918&0.936&0.888\\
$8\%$ & &  & 0.956&0.625&0.893&0.951&0.879&0.889&0.906&0.888&0.825&0.868 \\
$10\%$ & &  & 0.941&0.587&0.909&0.942&0.873&0.904&0.904&0.865&0.890&0.868 \\
\midrule
Null &  & & 0.975&0.321&0.873&0.864&0.947&0.791&0.879&0.953&0.915&0.835 \\
$2\%$ & & & 0.929&0.498&0.875&0.908&0.902&0.834&0.899&0.949&0.894&0.854\\
$4\%$ & & F1 & 0.950&0.468&0.914&0.907&0.904&0.834&0.874&0.924&0.910&0.854\\
$6\%$ & & & 0.934&0.518&0.897&0.917&0.892&0.808&0.887&0.934&0.921&0.856\\
$8\%$ & &  & 0.959&0.505&0.891&0.921&0.912&0.833&0.862&0.919&0.862&0.851 \\
$10\%$ & &  & 0.953&0.456&0.896&0.911&0.907&0.836&0.864&0.901&0.894&0.846 \\

\midrule
Null & & & 0.987&0.620&0.758&0.981&0.899&0.843&0.984&0.928&0.897&0.877 \\
$2\%$ & & & 0.976&0.738&0.697&0.977&0.851&0.847&0.978&0.924&0.908&0.877 \\
$4\%$ & \vistest & mAP & 0.984&0.730&0.758&0.977&0.879&0.842&0.980&0.903&0.918&0.886 \\
$6\%$ & & & 0.983&0.692&0.754&0.978&0.882&0.840&0.976&0.930&0.920&0.884 \\
$8\%$ & & & 0.977&0.690&0.699&0.978&0.865&0.840&0.976&0.904&0.899&0.870 \\
$10\%$ & & & 0.983&0.698&0.683&0.976&0.868&0.843&0.978&0.899&0.893&0.869\\
\midrule
Null & & & 0.993&0.571&0.932&0.952&0.932&0.841&0.946&0.942&0.953&0.896 \\
$2\%$ & & & 0.963&0.746&0.912&0.966&0.926&0.859&0.945&0.921&0.933&0.908 \\
$4\%$ &  & Precision & 0.990&0.739&0.926&0.966&0.940&0.863&0.954&0.908&0.926&0.913 \\
$6\%$ & & & 0.984&0.761&0.945&0.967&0.932&0.859&0.957&0.931&0.945&0.920 \\
$8\%$ & & & 0.985&0.788&0.930&0.965&0.928&0.867&0.951&0.924&0.948&0.921 \\
$10\%$ & & & 0.990&0.768&0.949&0.965&0.944&0.864&0.960&0.945&0.939&0.925\\

\midrule
Null & & & 0.990&0.722&0.775&0.984&0.909&0.888&0.987&0.942&0.913&0.901\\
$2\%$ & & & 0.983&0.792&0.715&0.981&0.864&0.891&0.985&0.940&0.934&0.898 \\
$4\%$ &  & Recall & 0.988&0.793&0.779&0.980&0.894&0.892&0.987&0.922&0.946&0.909\\
$6\%$ & & & 0.988&0.756&0.769&0.981&0.898&0.892&0.984&0.943&0.946&0.906\\
$8\%$ & & & 0.982&0.747&0.718&0.981&0.884&0.894&0.985&0.921&0.923&0.893\\
$10\%$ & & & 0.986&0.774&0.696&0.980&0.885&0.892&0.986&0.916&0.918&0.892\\

\midrule
Null & & & 0.991&0.638&0.846&0.967&0.921&0.864&0.966&0.942&0.932&0.896\\
$2\%$ & & & 0.972&0.767&0.800&0.973&0.893&0.875&0.965&0.930&0.933&0.901 \\
$4\%$ &  & F1 & 0.989&0.752&0.844&0.973&0.916&0.877&0.970&0.913&0.935&0.908\\
$6\%$ & & & 0.986&0.757&0.845&0.974&0.915&0.875&0.970&0.937&0.945&0.912\\
$8\%$ & & & 0.983&0.765&0.808&0.973&0.906&0.880&0.968&0.922&0.935&0.904\\
$10\%$ & & & 0.988&0.768&0.801&0.972&0.913&0.878&0.973&0.930&0.928&0.906\\

\bottomrule
\end{tabular}
\end{table}

\begin{figure*}[!t]
\centering
\subfloat{\includegraphics[width=\textwidth]{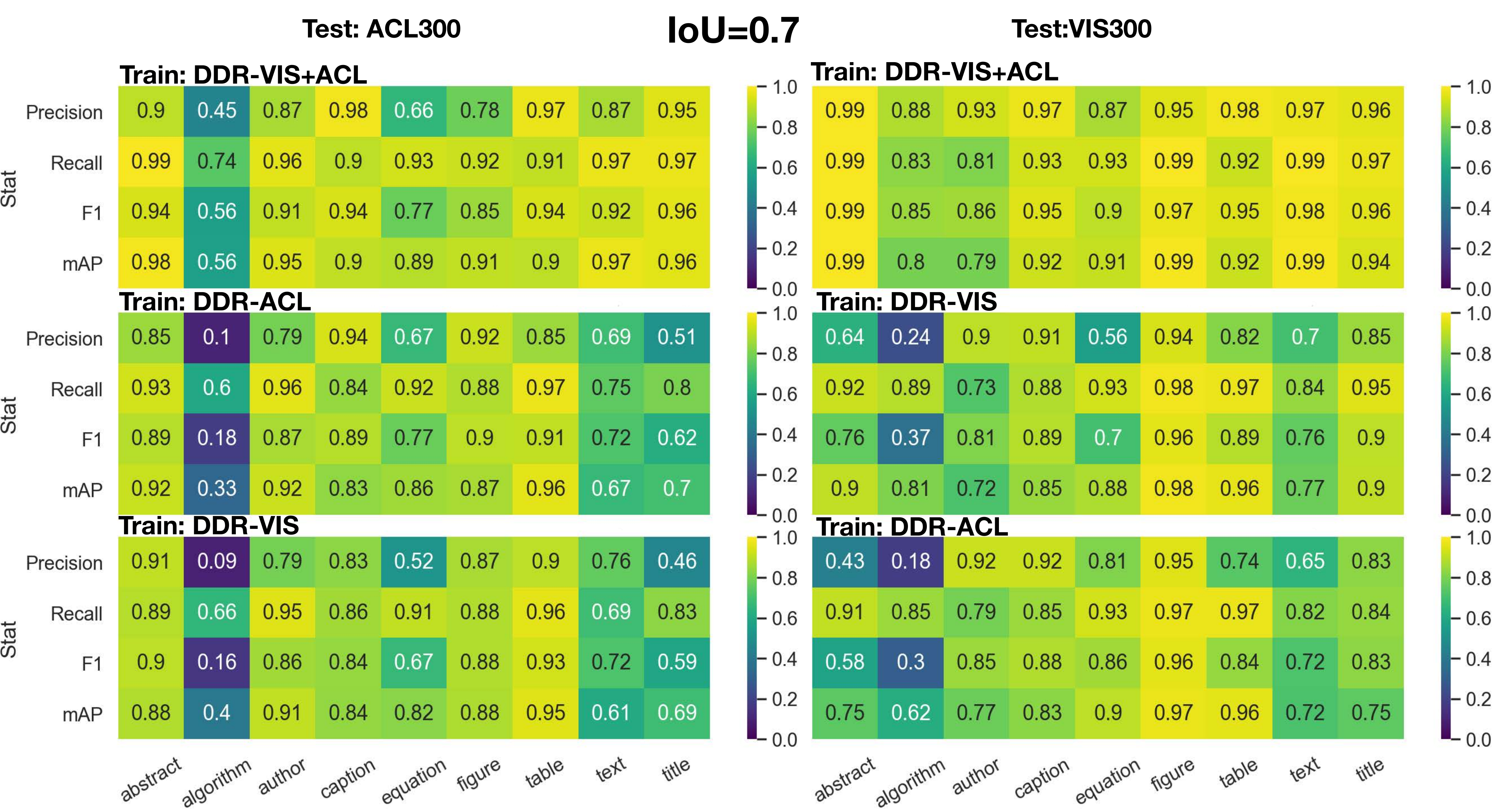}%
\label{fig:ddr0.7.1}}
\vspace{-10pt}
\subfloat{\includegraphics[width=\textwidth]{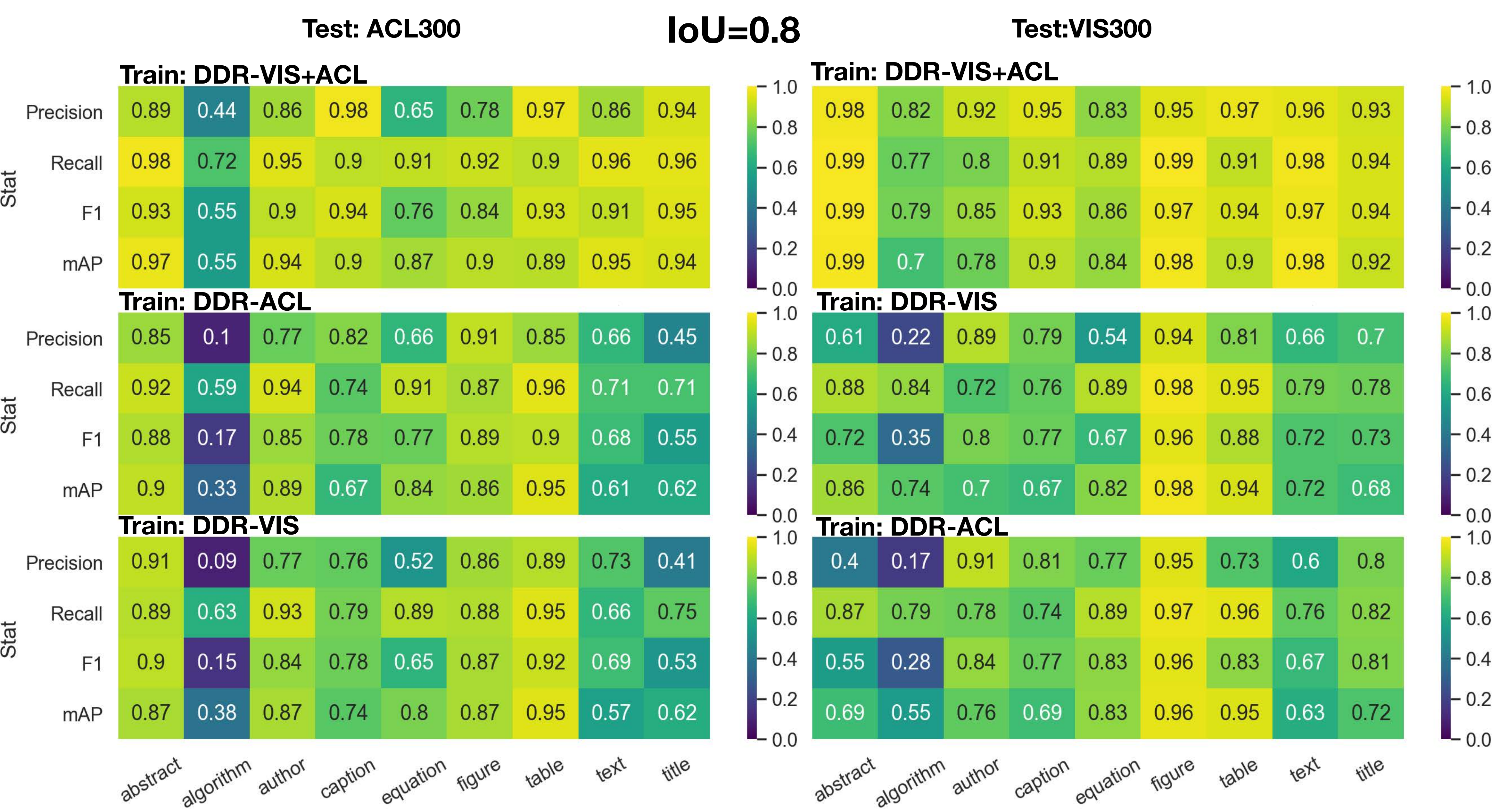}%
\label{fig:ddr0.8.1}}
\vspace{-10pt}
\subfloat{\includegraphics[width=\textwidth]{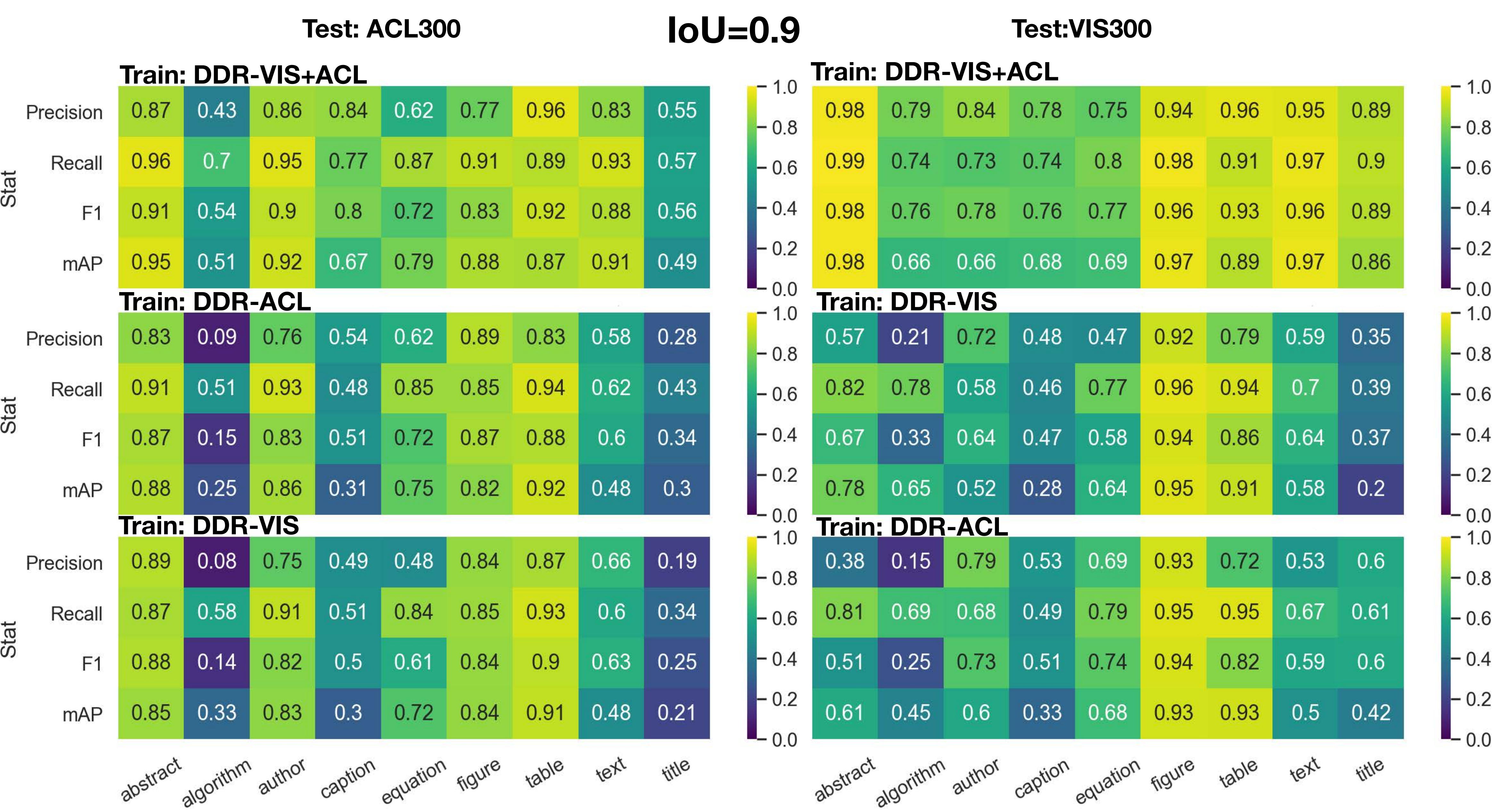}%
\label{fig:ddr0.9.1}}
\caption{DDR behavior results from six experiments in Study II.}
\label{fig:DDRmetrics}
\end{figure*}

\begin{figure*}[!t]
\centering
\subfloat{\includegraphics[width=\textwidth,frame]{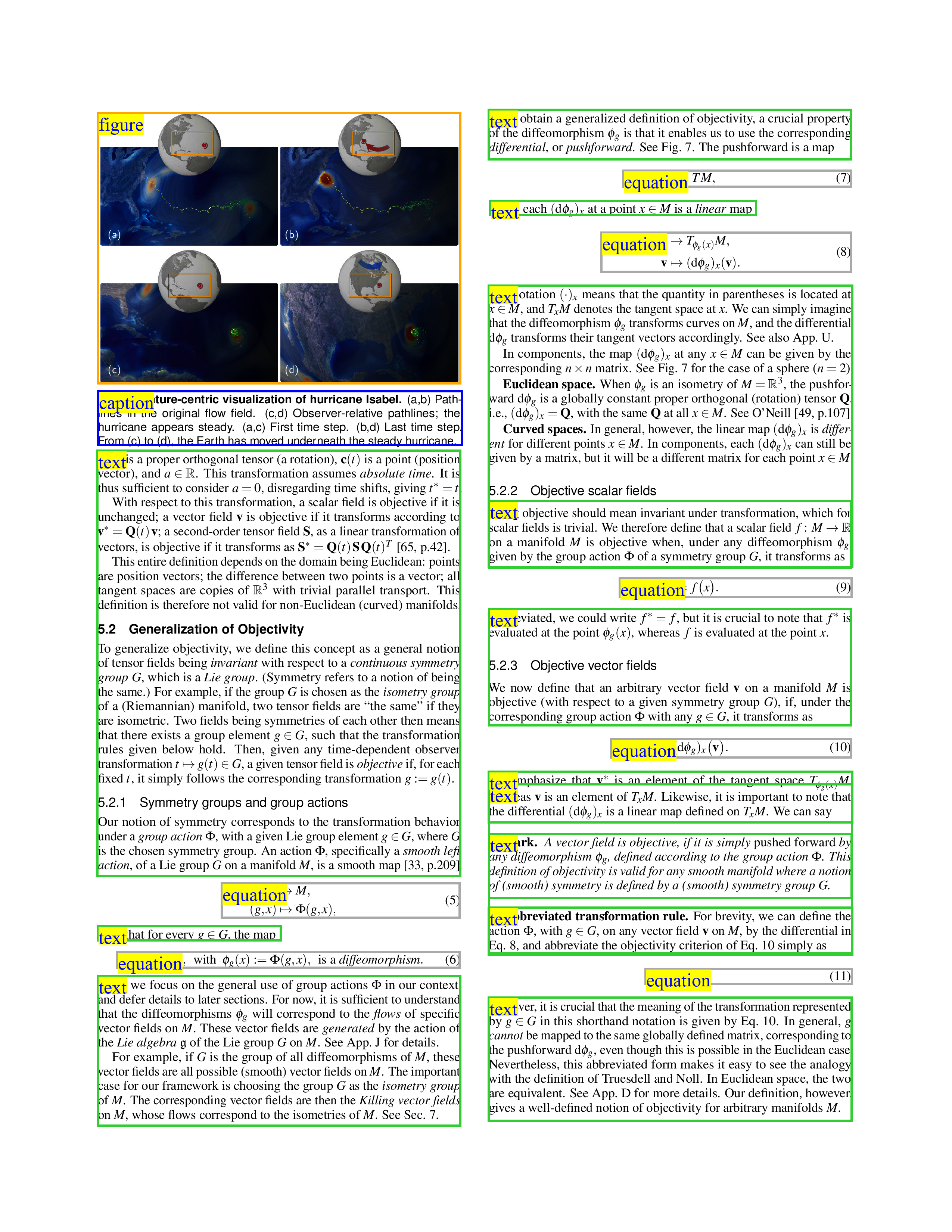}%
\label{fig:ddr0.7}}
\caption{Result sample: correctly labelled image with many equations and one figure/caption.}
\label{fig:pred1}
\end{figure*}

\begin{figure*}[!t]
\centering
\subfloat{\includegraphics[width=\textwidth,frame]{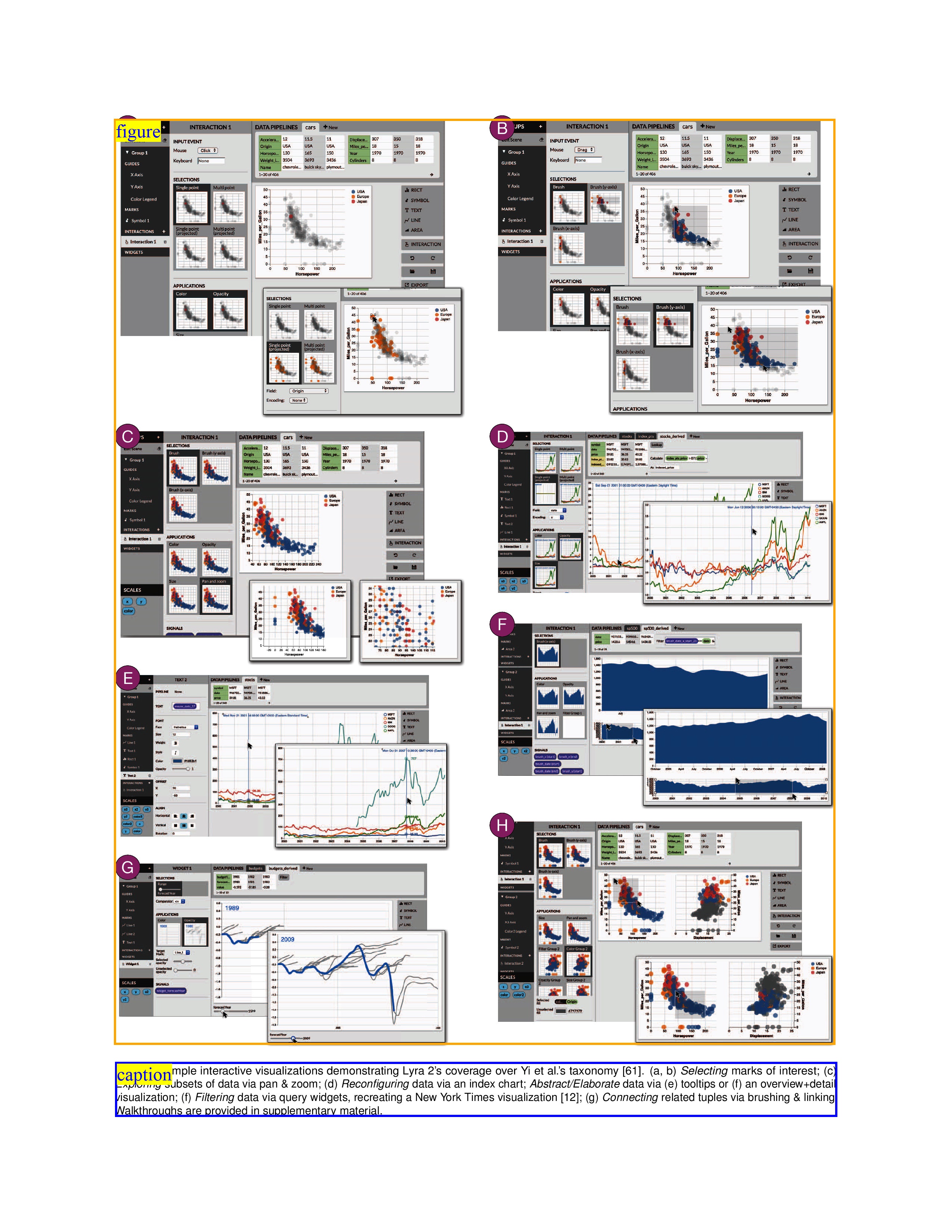}%
\label{fig:ddr0.8}}
\caption{Result sample: correctly labelled image that has many subimages.}
\label{fig:pred2}
\end{figure*}

\begin{figure*}[!t]
\centering
\subfloat{\includegraphics[width=\textwidth,frame]{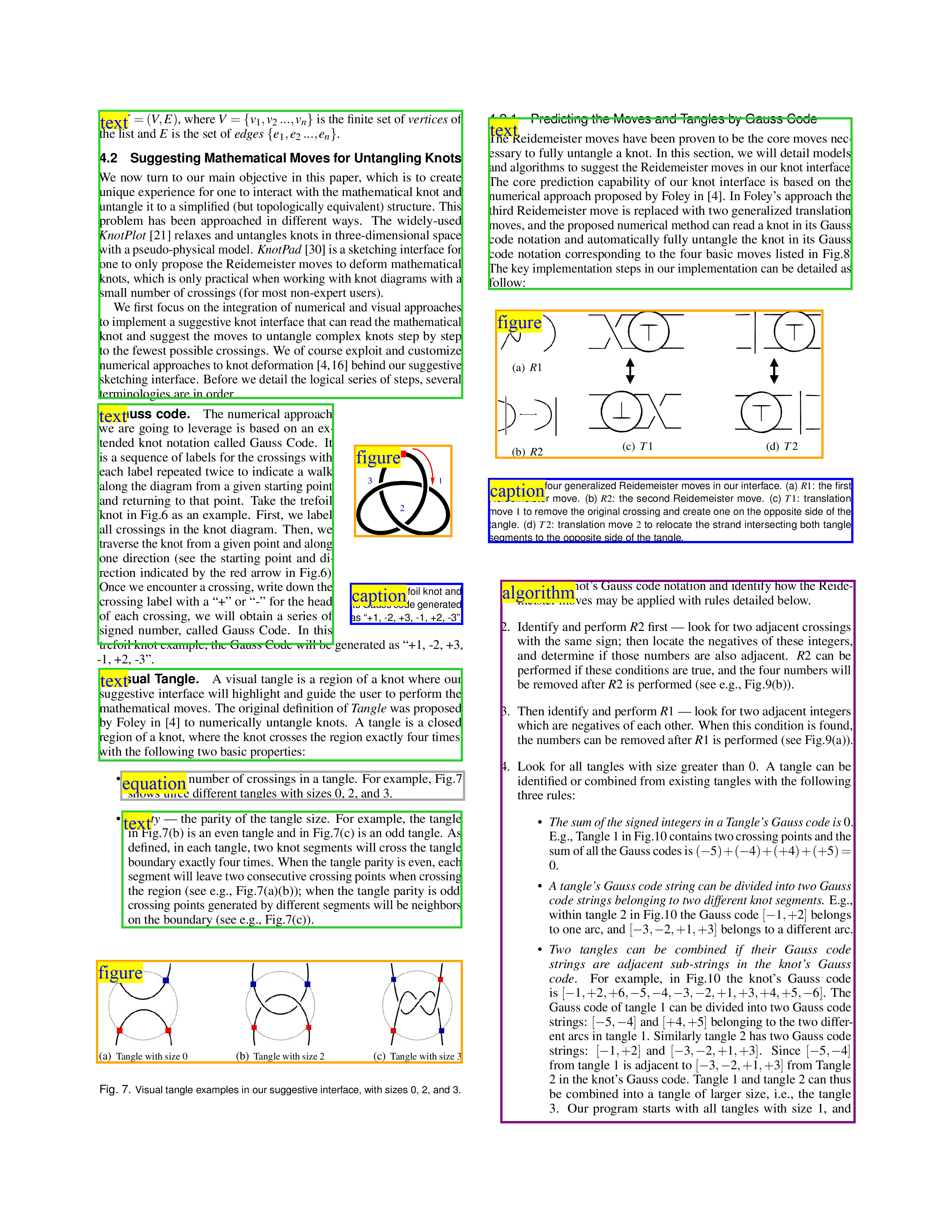}}%
\caption{Result sample: partially incorrectly labeled image: DRR recognized the small figure and its caption but labeled a bullet list as an algorithm and another as an equation. One caption is also missing. This result suggests that we may need to explicitly add ‘bullet list’ class to our training data.}
\label{fig:pred3}
\end{figure*}

\autoref{tab:DDR} shows the numerical values of \autoref{fig:DDR} 
in the main text for IoU of 0.8 for the six DDR experiments (trained on three styles and tested on \acltest and \vistest). 
\autoref{fig:DDRmetrics} presents the detection results for these
experiments for all IoUs of 0.7, 0.8, and 0.9, respectively. 
\autoref{fig:pred1}--\ref{fig:pred3} show some of the prediction results. 

We used four metrics (accuracy, recall, F1, and  mean average precision (mAP)) to evaluate
\cnns' performance in model comparisons, and the preferred ones are often chosen based on the object categories and goals of the experiment. For example, 
\begin{itemize}
\item
\textbf{Precision and recall.} \textit{Precision = true  positives $/$ (true positives + false positives))} and \textit{Recall = true positives / true positives + false negatives}.  
Precision helps when the cost of the false positives is high and is computed. Recall is often useful when the cost of false
negatives is high. 

    \item \textbf{mAP} is often preferred for visual object detection (here figures, algorithms, tables, equations), since 
    it provides an integral evaluation of matching between the ground-truth bounding boxes and the
predicted ones. The higher the score, the more accurate the model is for its task. 

\item \textbf{F1} is more frequently used in text detection. 
A F1 score represents an overall measure of a model's accuracy that combines precision and recall. 
A higher F1 means that 
the model generates few false positives
and few false negatives, and can identify
real class while keeping distraction low. Here,
\textit{F1 = 2 $\times$ (precision $\times$ recall) / ( precision + recall)}.
\end{itemize}

For simplicity, we used mAP scores in our own reports because they are comprehensive measures suitable to visual components of interest. However, in making comparisons with other studies for test on CS-150, we used the three other scores of precision, recall, and F1 because other studies did so. All scores are released for all study conditions in this work.

\section{Image Rights and Attribution}

The VIS30K~\cite{chen2020vis30k} dataset comprises all the images published at IEEE visualization conferences in each year, rather than just a few samples. All image files are copyrighted and for most the copyright is owned by IEEE. The dataset was released on IEEE Data Port~\cite{Chen:2021:VCF}. We thank IEEE for dedicating tools like this to support the Open Science Movement. All ACL papers are from the ACL Anthology website.

\end{document}